\documentclass{article}

\PassOptionsToPackage{numbers, sort&compress}{natbib}

 \usepackage[main, final]{neurips_2026}

\usepackage[utf8]{inputenc}
\usepackage[T1]{fontenc}
\usepackage{hyperref}
\usepackage{url}
\usepackage{booktabs}
\usepackage{amsfonts}
\usepackage{nicefrac}
\usepackage{microtype}
\usepackage{xcolor}
\usepackage{amsmath}
\usepackage{amssymb}
\usepackage{graphicx}
\usepackage{algorithm}
\usepackage{algpseudocode}
\usepackage{placeins}
\usepackage{cleveref}
\usepackage{titletoc}
\usepackage{enumitem}
\Crefname{section}{Sec.}{Secs.}
\crefname{section}{Sec.}{Secs.}
\usepackage[subtle]{savetrees}

\newcommand{\tabtitle}[1]{\textit{#1}}

\title{As the Story Unfolds: Watching a Film and \\ Identifying Characters as a Human Does
}

\author{%
  Zhongrui Gui \textsuperscript{1} \quad
  Junyu Xie\textsuperscript{1} \quad
  Tengda Han\textsuperscript{1} \quad
  Weidi Xie\textsuperscript{1, 2} \quad
  Andrew Zisserman\textsuperscript{1} \\
  \textsuperscript{1}Visual Geometry Group, Department of Engineering Science, University of Oxford, UK \\
  \textsuperscript{2}School of Artificial Intelligence (SAI), Shanghai Jiao Tong University, China \\
  \texttt{\{zhongrui,jyx,htd,weidi,az\}@robots.ox.ac.uk}
}

\begin{document}

\maketitle

\begin{figure}[th]
    \centering
    \vspace{-1mm}
    \includegraphics[width=\linewidth]{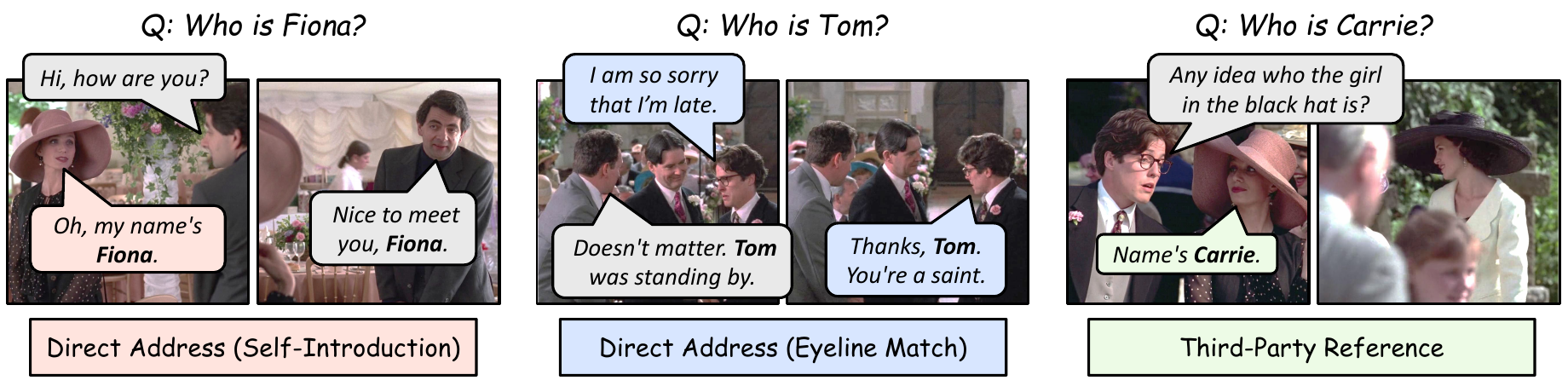}
    \vspace{-5.5mm}
    \caption{\textbf{Character naming task.}
    We associate character names to their facial appearances in movies, mirroring how viewers infer who is who while watching. Characters may be introduced through diverse cues, including self-introduction, direct address, and third-party reference. We show three representative examples from \textit{Four Weddings and a Funeral} (1994). Correctly linking a character to their name is central to movie understanding, enabling coherent character tracking across scenes as the story unfolds.}
    \label{fig:teaser}
    \vspace{-1mm}
\end{figure}

\begin{abstract}
Films introduce characters incrementally as they progress. We study the problem of identifying principal characters and assigning their names directly from the movie itself, without using external cast lists, actor photographs, or pre-built character banks. We introduce the character naming task: given a name mentioned in dialogue and the associated video context, the model must determine which visible character bears that name. This task is challenging because name resolution often depends on cinematic cues such as shot–reverse-shot structure, self-introduction, direct address, third-party reference, gaze, and temporal continuity. We provide a dataset and annotations for evaluating character naming and first character appearances, develop a video-and-dialogue model for the task, and compare it with proprietary multimodal large language models. We further introduce a progressive framework for building a character bank as the film unfolds. Together, these contributions support online, human-like movie understanding in which character identities are inferred from the film itself.
Website: \url{https://www.robots.ox.ac.uk/~vgg/research/StreamCast/}.
\end{abstract}

\section{Introduction}
The objective of this paper is to watch a film as a human would -- ingesting the film frame-by-frame in an online mode,
and identifying the principal characters and their names {\em as the story unfolds}. Traditional films, such as a romantic comedy,
consist of characters and their story arcs. In order for the viewer to quickly understand who these principal characters are, the
film must introduce and name them towards the start of the film. This is achieved by a combination of cinematic devices (such as the camera following the person, and close-ups) and using the dialogue or other information to name them, as illustrated in Fig.~\ref{fig:teaser}.

Obtaining character names explicitly only from the film is in stark contrast to the methods employed recently in the literature for tasks like generating audio descriptions and transcripts~\cite{pini2019m, han2023autoad, han2023autoadii, han2024autoadiii, xie2024autoad, xie2025shot, gui2025character}. In these works, a {\em character bank} is first constructed using a cast list and actor photos from IMDB, Fandom and other online sources. The character bank is then used to label the principal characters throughout the film using face recognition (or voice recognition for animated films~\cite{gui2025character}).

To operationalize our goal, we first investigate how the viewer is
presented with the name of a character as they watch the film --
almost always it is from the dialogue, and very occasionally from
other sources, such as name badges. This enables the {\em character naming}
task to be automated and evaluated -- given a name in
the dialogue, the associated shots can be used to ask the question:
which of the people appearing in this clip has this name? To this end,
the first contribution of this paper is to build a dataset from
MovieNet~\cite{huang2020movienet} and Movie101~\cite{yue2023movie101} films that can be used to train and evaluate this
task. We develop a method to solve this task, using the video and
dialogue, and compare its performance to proprietary Multimodal LLMs
(MLLMs), such as Gemini~\cite{gemini3}. This is a challenging task, as correct
inference often requires an understanding of film editing style (as
the person addressed by name may be shown in a different shot) and
also character gaze and eyeline (to know who they are looking at when
saying a name).

This establishes that characters can be named in films, using the information presented by the screenwriter and director. Often, the name of a principal character is presented multiple times across different dialogues throughout the film, providing opportunities for error correction (by both human and machine) if the first presentation had been missed or incorrectly assigned.
Once the principal characters are named (possibly multiple times), they can be recognized throughout the rest of the movie, and retrospectively named from their first appearance (using human or machine memory).
To this end, we introduce two further tasks: (i) annotating the movies with the first appearance of the principal characters (which may well be before they are named in the dialogue); and (ii) progressively building up a {\em character bank} as the film progresses. The character bank records the appearance, voice and names of the principal characters (as in previous work). We provide annotations to evaluate the character first appearance over a number of movies. For example, in the movie \textit{Juno}, the protagonist \textit{Juno} is introduced in the first shot of the film, and appears in almost every scene, but is not named until $7.2$ minutes into the film.

In summary, we make the following four contributions:
(i) we introduce the new task of \emph{character naming} from dialogue alone, a key enabling step for watching and understanding a film as a human does; (ii) we provide the Movie Character Naming (MCN) dataset with $23,792$ instances of multiple-choice character naming over $407$ movies, and also $20$ movies with the first appearance of the principal characters annotated; (iii)
we develop a character naming model using open-source MLLMs, and compare its performance to proprietary MLLMs, analysing what is difficult to solve; and (iv)
we introduce the first automatic method to progressively build up a character bank for a feature-length movie. These capabilities put in place the foundation for understanding character arcs and, ultimately, the story of a film.
The dataset and models are available at: \url{https://www.robots.ox.ac.uk/~vgg/research/StreamCast/}.

\section{Movie Character Naming (MCN) Dataset}
In this section, we introduce the Movie Character Naming (MCN) Dataset, which provides annotations for two complementary tasks: \emph{Character Naming} and \emph{First Occurrence}. The \emph{Character Naming} annotations consist of video segments in which character names are uttered in the dialogue, each paired with ground-truth name-to-track mappings. The \emph{First Occurrence}
annotations, by contrast, are designed to evaluate the streaming pipeline for character bank construction, and provide two (possibly coinciding) timestamps per principal character: the moment they first appear on screen, and the moment their name is first revealed.

\begin{figure}[t]
    \centering
    \includegraphics[width=0.95\linewidth]{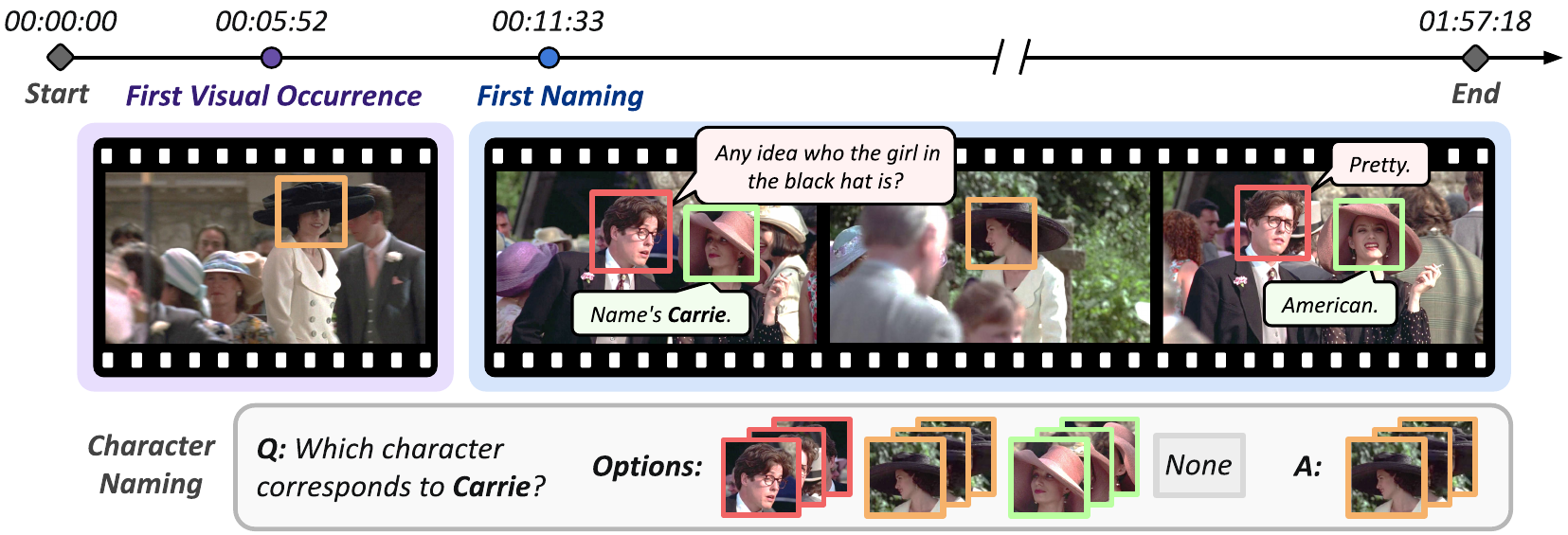}
    \vspace{-2.5mm}
    \caption{\textbf{Example annotations from the MCN dataset.}
    The MCN dataset provides manually annotated timestamps for each character’s first visual occurrence and first naming, along with ground-truth name-to-face correspondences for the \emph{Character Naming} task.
}
    \label{fig:sample}
    \vspace{-1.0mm}
\end{figure}

Fig.~\ref{fig:sample} shows a representative MCN sample for the character \textit{Carrie} from \textit{Four Weddings and a Funeral}. The movie timeline is annotated with \emph{Character Naming} and \emph{First Occurrence} labels. Additional examples are provided in Appendix~\ref{supp:dataset}. In the following, we first describe how the dataset is annotated given character tracks and dialogue. The implementation details for obtaining the character face-tracks and dialogue are given in Sec.~\ref{preprocessing}.

\subsection{Data Collection and Curation}
We build MCN on top of MovieNet~\citep{huang2020movienet} and Movie101~\cite{yue2023movie101}, which already provide ground-truth mappings between character names and body bounding-box coordinates.

\textbf{Annotations for Character Naming (CN).} We construct the associations between character names and face tracks in three steps. \emph{First,} since MovieNet and Movie101 already provide ground-truth name--body associations, we align character body boxes with their face tracks in space and time, yielding name--face associations for \emph{all} characters appearing on screen. \emph{Second,} we keep only those associations whose corresponding names are also mentioned in the dialogue, since these are the only ones inferable from the audio--visual signal we use. \emph{Third,} we handle the case where a name is mentioned in the dialogue but the corresponding face track is unavailable (e.g., when the character is referred to in absentia). In this case, we label the association as \texttt{none}, since no in-clip cues support the assignment. As a result, we provide the final name--face associations as ground-truth labels, along with all character-level face tracks in the video clip as candidate options, to construct a multiple-choice instance.

\textbf{Annotations for First Occurrence (FO).} For every character in the movie, we manually annotate two timestamps: the \emph{first visual occurrence}, i.e.\ the moment the character first appears on screen, and the \emph{first naming}, i.e.\ the moment at which a human viewer can first associate the character's visual appearance with their name based solely on the preceding content.

\textbf{Quality Control.} To ensure a high-quality test set, we manually inspect every test sample and correct errors in face-track clustering and name labelling. We then run state-of-the-art proprietary models on the curated set and examine cases where their outputs disagree with the annotation, fixing any remaining errors (or confirming the annotation is correct but the model fails).

\subsection{Dataset Statistics}
The statistics of both annotation types in the MCN dataset are summarised in Tab.~\ref{tab:dataset_stats}. For the CN task, the training splits comprise clips from $377$ training movies ($108.13$ hours), covering $1{,}853$ characters and $22{,}210$ name-to-track questions, of which $21{,}150$ have non-\texttt{none} answers. The training movies are disjoint from those in both the CN and FO test sets, ensuring no movie-level leakage across splits.
The CN test split contains clips from $30$ movies ($6.51$ hours), covering $226$ characters and $1{,}582$ manually verified questions, of which $1{,}399$ have non-\texttt{none} answers. In particular, the CN questions in the test set can be broken down by evidence type: approximately $12\%$ involve self-introduction, $70\%$ involve direct address by other characters, $16\%$ involve third-party reference, and $2\%$ require reasoning from textual cues in the movie, such as name badges.

The FO annotations form a test set of $20$ movies ($39.98$ hours) with $287$ annotated characters, yielding $574$ timestamp annotations in total. Each character has two annotations, corresponding to the first visual occurrence and the first naming. Together, the MCN dataset spans a diverse range of genres, casts, and narrative styles, ensuring that our evaluation captures performance across varied cinematic settings.

\begin{table}[t]
    \centering
    \caption{\textbf{Statistics of the Character Naming (CN) and First Occurrence (FO) annotations.} Each question corresponds to a name-to-track mapping query or a first-occurrence timestamp annotation.}
    \vspace{-1.5mm}
    \setlength{\tabcolsep}{8pt}
    \resizebox{0.75\linewidth}{!}{%
    \begin{tabular}{cccccc}
        \toprule
        \tabtitle{Task} & \tabtitle{Split} & \tabtitle{\# Movies} & \tabtitle{Duration (h)} & \tabtitle{\# Characters} & \tabtitle{\# Questions} \\
        \midrule
        CN  & Train & $377$ & $108.13$ & $1853$ & $22210 / 21150$ \\
        CN  & Test  & $30$ & $6.51$ & $226$ & $1582 / 1399$ \\
        FO  & Test  & $20$ & $39.98$ & $287$ & $574$ \\
        \bottomrule
    \end{tabular}}
    \label{tab:dataset_stats}
    \vspace{-1.5mm}
\end{table}

\subsection{Preprocessing -- Character-Level Face-Tracks and Associated Speech}
\label{preprocessing}
To obtain multiple-choice instances for \emph{Character Naming}, we need to extract character-level tracks and speech from video clips around name occurrences in the dialogue, therefore enabling us to ask models to select one of them for each query name. This involves extracting a face-track cluster $f_i$ for each character and linking it with the corresponding voice-track cluster $v_j$. Note, this pre-processing is done at the clip level when annotating the dataset, but results are also made available over the entire movie for downstream tasks (e.g., character bank construction).

\textbf{Face Track Extraction.} For each recurring character $i$, we detect and track faces across shots and group the tracks into a character-level face-track cluster $f_i$. The detailed procedure is given in Appendix~\ref{supp:clustering}.

\textbf{Linking Voices to Face Tracks.}
For the clean voice segments in the video, we first apply speaker diarization to group them into voice clusters, with each cluster denoted $v_j$. To match each voice cluster to an existing face track (i.e.\ to identify who is speaking on screen), we apply TalkNet~\citep{tao2021someone} to obtain synchronization scores $s_{i,j}$ between faces $f_i$ and voices $v_j$. Each voice cluster is then assigned to the face-track cluster with the highest synchronization score, yielding complete character identities with paired faces and voices.

As a further preprocessing step, we transcribe the voice tracks into textual dialogue with Parakeet~\citep{rekesh2023fast}. Since each voice track is linked to a face cluster, this naturally yields visually grounded subtitles.
Finally, the character names $n$ emerge as a natural by-product of subtitle extraction. We retain clips whose subtitles contain at least one character name, passing them to the subsequent character naming step.

\section{Assigning Names to Characters}
\label{si}

The task of assigning names to characters, i.e., Character Naming (CN),
is performed naturally by humans while watching a film, but is not trivial for deep learning systems owing to multiple challenges: first, a character may be addressed directly by name, or referred to in the third person while off-screen, or revealed only through what others say about them; second, who is being addressed may only be indicated by gaze or by the editing structure, for example, as one character says a name, and the named character then reacts in the subsequent shot. We therefore treat CN as a new multimodal task that requires joint reasoning over textual cues (e.g., conversational logic, vocative address) and visual cues (e.g., eye contact, on-screen presence).

Formally, given a video clip $V$ containing a set of face-track clusters $\{f_i\}$ (each optionally linked to a voice cluster $v_j$, as described in Sec.~\ref{preprocessing}) and a candidate name $n$ extracted from the dialogue, the Character Naming (CN) task is to predict an assignment $\psi(n) \in \{f_i\} \cup \{\varnothing\}$. Here, $\psi(n) = f_i$ indicates that the name $n$ refers to the character represented by face-track cluster $f_i$, while $\psi(n) = \varnothing$ indicates that none of the visible characters corresponds to $n$ (e.g., the named character is off-screen).

In this section, we first describe how we operationalize Character Naming for MLLMs through visual prompting alone (Sec.~\ref{sec:prompting}). We then introduce a fine-tuning recipe that improves the MLLMs' reasoning on this task, enabling more reliable assignments of character names (Sec.~\ref{sec:finetuning}).

\subsection{MLLM-Based Character Naming}
\label{sec:prompting}
To adapt MLLMs for character naming in a zero-shot manner, we use the following prompting strategies.

\noindent\textbf{Prompting the MLLM with Visual Identities.} Rather than providing the MLLM with bounding-box coordinates of the face tracks, we adopt the color-based visual prompting scheme of~\citet{xie2024autoad}. Each face track is assigned a unique color, and its bounding boxes are drawn in that color on the raw video frames. The boxes are then referred to by color in the text prompt, allowing the MLLM to associate identities with their spatial locations.

\noindent\textbf{Prompting the MLLM with Visually Grounded Subtitles.} Subtitles are embedded directly into the text prompt, with each speaker referred to by the color tag of their associated face track. We additionally include timestamps to enable more temporally accurate inference. For example, a line spoken by the character with blue face bounding boxes would be formatted as~\texttt{<693.2s>~[BLUE]~\textit{``Any idea who the girl in the black hat is?''}~<695.0s>}.

\noindent\textbf{Prompting the MLLM for Character Naming.} Given the visual identities and visually grounded subtitles, we directly ask the MLLM to match each character name from the subtitles with a color, e.g., ``\texttt{\textit{Which color corresponds to John Constantine?}}'' The color code thus bridges each visual identity's spatial position to its corresponding name.

\subsection{Fine-tuning the MLLM for Character Naming}
\label{sec:finetuning}
Following the prompting strategies described above, we adopt a two-stage R1-style fine-tuning recipe~\citep{guo2025deepseek} to further improve the naming accuracy.

\noindent\textbf{Stage I: Supervised Fine-tuning (SFT).} The first stage is a cold start that supervises the MLLM with synthetic reasoning traces. These traces are generated by a teacher MLLM, which is given the video frames, visually grounded subtitles, and ground-truth name-to-color mapping, and asked to produce a reasoning trace that leads to the correct answer. This teaches the student MLLM to follow the desired output format (i.e. a reasoning trace followed by the answer) before the reinforcement learning stage.

\noindent\textbf{Stage II: Reinforcement Learning.} The second stage applies GRPO~\citep{guo2025deepseek} with a composite reward $R = r_f + r_a$, where $r_f$ rewards adherence to the expected response structure and $r_a$ measures the correctness of the predicted name--face pairs. Letting $\hat{i}(n)$ and $i^{*}(n)$ denote the face track assigned to name $n$ under the prediction $\hat{\psi}$ and the ground truth $\psi^{*}$, we set $r_a = \sum_{n \in N} \mathbf{1}[\hat{i}(n) = i^{*}(n)]$. The GRPO advantage and policy update follow the standard formulation~\citep{guo2025deepseek}, summarized in Appendix~\ref{supp:grpo}.

A naive application of GRPO to this task suffers from a subtle shortcut: the model can latch onto color-specific cues. For instance, it may always associate the protagonist with ``red'' simply because ``red'' is the most frequently used color in the training distribution. We address this with a simple but effective \emph{color-permutation augmentation}: during RL training, we randomly permute the color assignment for each sample. The ground-truth name--face mapping is invariant under such permutations, so consistent rewards across permuted variants are achievable only by reasoning about identity itself. This regularizes the policy toward color-agnostic reasoning and, empirically, accounts for a significant share of our gains over the cold-start baseline.

\section{Streaming Pipeline for Character Bank Construction}
Given a full movie, our goal is to construct a character bank in an online manner, taking the shot as the elementary unit. The \emph{character bank} $\mathcal{B}_t$ serves as a memory of all characters observed so far. At each shot $t$, it aggregates the corresponding character information as triplets $(f, n, v)$, where $f$ is a face-track cluster, $n$ a character name, and $v$ a voice cluster.

\begin{figure}[t]
    \centering
    \includegraphics[width=\linewidth]{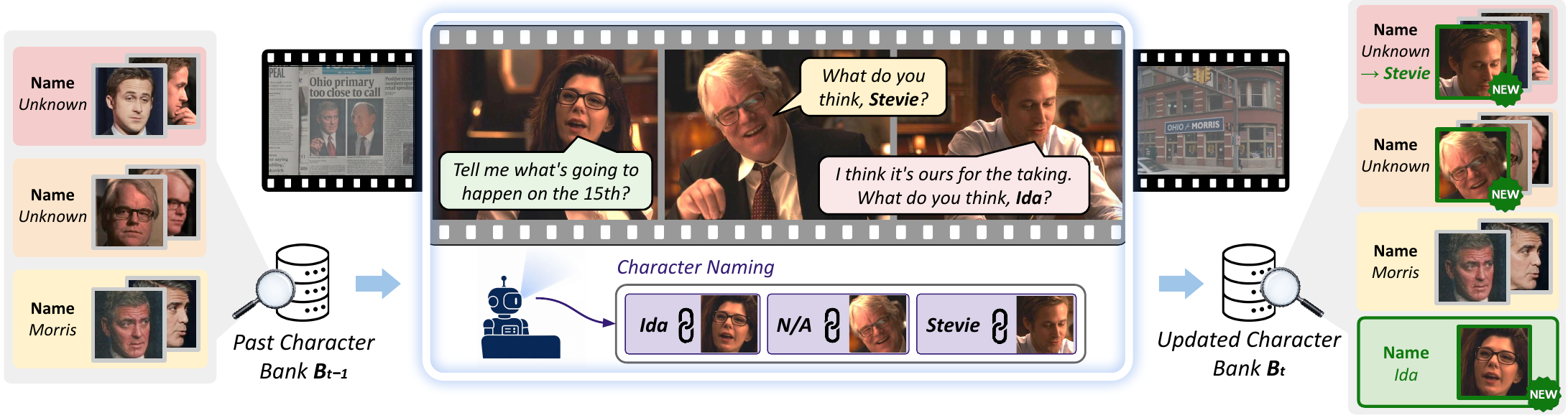}
     \vspace{-6.5mm}
    \caption{\textbf{Streaming pipeline for character bank construction.}
   As new shots arrive, we extract observations (character tracks, subtitles, and mentioned names) and reason jointly over them to link face tracks to names. Each step incrementally updates the bank through three operations: \emph{naming} an existing cluster (\textit{Stevie}, first cluster), \emph{enriching} a cluster with additional face exemplars (second cluster), and \emph{instantiating} a new entry for a previously unseen character (fourth cluster). The sample is drawn from \textit{The Ides of March} (2011). }
    \label{fig:construction}
    \vspace{-2mm}
\end{figure}

Formally, the bank is updated auto-regressively (as illustrated in Fig.~\ref{fig:construction}):
\begin{equation}
\label{eq:mapping}
    \mathcal{B}_t = \phi_{\mathcal{B}}(\mathcal{B}_{t-1}, \psi(V_t)),
\end{equation}
where $V_t$ denotes the video clip at shot $t$, and $\psi$ is the MLLM fine-tuned for CN, which outputs character names and corresponding face tracks and voices $\{(n, f, v)\}$. We also retain identities with no assigned name, i.e.\ $\{(\varnothing, f, v)\}$, mirroring how a human viewer can visually remember a character even before they are introduced by name. These triplets are then merged into the previous bank by the update operator $\phi_{\mathcal{B}}$.

Overall, this streaming pipeline maintains and auto-regressively updates the character bank $\mathcal{B}$.
We detail the bank update process in Sec.~\ref{sec:update} and introduce an error-correction mechanism in Sec.~\ref{sec:error_correction} that keeps the update robust to error accumulation over multiple steps.

\subsection{Character Bank Update}
\label{sec:update}
Given the names and corresponding face and voice tracks $\{(n, f, v)\}$ extracted from a new shot (with $n$ possibly empty), we merge them into the existing entries of the previous bank in two steps.

\textbf{Track Assignment.} We treat face tracks as the primary signal and assign each $f_i$ to an existing face cluster in the bank. We extract the corresponding face embedding $e_i$ and find the closest cluster by distance to its embedding centroid $\bar{e}_b$ as in Eq.~\ref{eq:update}. A smaller distance indicates greater similarity between the face embedding $e_i$ and the cluster centroid $\bar{e}_b$.
\begin{equation}
\label{eq:update}
    b^{*} = \arg\min_{b \in \mathcal{B}_{t-1}} d_{\cos}(e_i, \bar{e}_b).
\end{equation}
If $d_{\cos}(e_i, \bar{e}_{b^*}) < \tau_f$, the track is assigned to entry $b^{*}$. Otherwise, we fall back to voice as the secondary signal: if the shot's speaker embedding $u_i$ matches an entry's voice centroid $\bar{u}_b$ under a tighter distance threshold $\tau_v < \tau_f$, the track is assigned there. If neither threshold is satisfied, a new entry is instantiated in the bank. Upon assignment, the matched centroids are updated.

\textbf{Name Binding.} We then bind a name to each entry in the character bank. For each entry, we maintain a histogram over the names associated with its face tracks across all observed shots, and use the majority-voted name as its canonical name. This handles aliases and nicknames; for example, in \textit{Juno}, the character \textit{Juno} may be addressed as \textit{Mama Bear}, \textit{Home Skillet}, or \textit{General MacGuff}. Majority voting helps recover the most representative name for each character. It also improves robustness to noisy predictions from the character naming module, since occasional erroneous name assignments are unlikely to dominate the histogram.

\noindent\textbf{Retrospective Assignment} is a special case of name binding: when the CN module assigns a name to a previously unnamed face-track cluster, the new name is added to that cluster's histogram with a count of one, and the canonical name is updated accordingly. Thus, clusters that are initially unnamed can still acquire names once sufficient naming evidence becomes available in later shots.

\subsection{Error Correction during Character Bank Construction}
\label{sec:error_correction}
The Markov structure of the update process $\phi_{\mathcal{B}}$ makes it susceptible to error accumulation from upstream components (e.g.\ transcription, active-speaker detection, and identity grounding). To mitigate this, we introduce three error correction mechanisms deployed along the streaming chain. First, we use an MLLM to filter out noisy or ambiguous face tracks before clustering, ensuring high-quality embeddings. Second, we use an LLM to filter out query names from the subtitles that are unlikely to refer to on-screen characters. Third, we resolve cross-entry name conflicts after each shot using a greedy voting scheme that enforces one-to-one name assignment. Together, these mechanisms address distinct error modes and stabilize the evolution of $\mathcal{B}_t$ without revisiting past observations. We refer the reader to Appendix~\ref{supp:error_corr} for further details.

\section{Experiments}

This section presents our evaluation. We first introduce the metrics (Sec.~\ref{subsec:metrics}), then evaluate each task. Sec.~\ref{exp_si} evaluates our character naming (CN) module on the MCN test set against state-of-the-art open- and closed-source models. Sec.~\ref{exp_linkage} evaluates the full streaming character bank pipeline, measuring how well our method tracks each character's first visual appearance and the first moment it can be linked to a name, and compares the final character bank against the IMDB ground truth.
We ablate key design choices, including the two-stage fine-tuning strategy and data augmentation for CN, and the error correction mechanisms for the character bank. Implementation details and ablations are provided in Appendix~\ref{supp:implementation} and Appendix~\ref{supp:ablation}.

\subsection{Evaluation Metrics}
\label{subsec:metrics}
\textbf{Character Naming Accuracy.} We evaluate for two settings. \emph{Overall accuracy} is computed over all name queries, including those labelled \texttt{none} when the mentioned character cannot be linked to any on-screen face track. \emph{Known accuracy} is computed only over queries whose ground-truth label is a valid face track, thereby examining the model's ability to correctly ground on-screen characters.

\textbf{Character Bank Mean Temporal IoU.}
To evaluate the character bank streaming pipeline, we propose a \emph{Mean Temporal IoU} (mTIoU) metric that measures how closely the bank tracks ground-truth human knowledge of each character over time.

For a movie of duration $T$ and character $c$, let $H_c \subseteq [0, T]$ denote the interval during which a human viewer is aware of $c$, and $P_c \subseteq [0, T]$ the interval during which the model correctly identifies $c$ in the bank. We define $\mathrm{TIoU}(c) = |H_c \cap P_c|/|H_c \cup P_c|$, which penalizes both \emph{late} identification and \emph{incorrect} identification (e.g.\ an entry later overwritten with a wrong name). To reflect narrative importance, we weight each character by their subtitle mention frequency $w_c$ and report $\mathrm{mTIoU} = \sum_c w_c \, \mathrm{TIoU}(c) / \sum_c w_c$.

We instantiate two variants from the MCN timestamps: the \emph{visual} variant $\mathrm{mTIoU}_v$ uses $H_c = [t_c^v, T]$ with $t_c^v$ the first visual occurrence of $c$, and the \emph{name} variant $\mathrm{mTIoU}_n$ uses $H_c = [t_c^n, T]$ with $t_c^n$ the first naming timestamp (when a human can first link a face to a name).

\textbf{Principal Character Recall (PCR@K).} To evaluate the quality of the final character bank, we propose \emph{Principal Character Recall} at $K$ (PCR@$K$), measuring how reliably the pipeline recovers a movie's most important characters with both correct names and correct visual exemplars.

For each movie, we take the top-$K$ principals from IMDB, ranked by their narrative importance. A ground-truth principal $c$ is \emph{recovered} iff some single entry $b \in \mathcal{B}$ satisfies both of the following conditions: (i) the canonical name of $b$ phonetically matches the IMDB name of $c$, and (ii) the distance between its face centroid $\bar{e}_b$ and the IMDB reference embedding $\bar{e}_c^{*}$ is below the threshold $\tau_{\text{eval}}$.
We report per-movie recall as the fraction of recovered principals and average across movies.

\subsection{Results on Character Naming}
We evaluate Character Naming (CN) with three sets of methods. First, we compare against several baselines, including a random baseline that uniformly selects from the candidate colors, non-proprietary open-source MLLMs (Qwen3VL-4B-Thinking, Qwen3VL-8B-Instruct, and Qwen3.5-9B~\cite{bai2025qwen3}), and a naive fine-tuning baseline obtained by directly fine-tuning Qwen3VL-8B-Instruct on the MCN training set. Second, we report the result of our CN model, which is obtained by two-stage R1-style fine-tuning of Qwen3VL-4B-Thinking on the MCN training set. Third, we compare against state-of-the-art proprietary MLLMs: Gemini-3-Flash-Preview and Gemini-3.1-Pro-Preview~\cite{gemini3}. All models receive the same inputs with a visually prompted video clip, visually grounded subtitles, and queried character names, and are asked to output the corresponding box colors.

\label{exp_si}

\begin{table}[t]
    \centering
    \caption{\textbf{Character naming accuracy on the MCN test set.} We report overall accuracy (including \texttt{none} labels) and known accuracy (restricted to queries with a valid face track). Our fine-tuned Qwen3VL-4B-Thinking closes most of the gap to proprietary frontier models despite being substantially smaller in terms of model size.}
    \vspace{-1mm}
    \setlength{\tabcolsep}{6pt}
    \resizebox{0.83\linewidth}{!}{%
    \begin{tabular}{lcccc}
        \toprule
        \tabtitle{Model} & \tabtitle{Propriet.} & \tabtitle{Fine-Tuned} & \tabtitle{Overall Acc.} & \tabtitle{Known Acc.} \\
        \midrule
        Random Baseline                 & $\times$ & $\times$   & $29.90$ & $33.74$ \\
        \midrule
        Gemini-3-Flash-Preview          & \checkmark & $\times$   & $79.57$ & $86.62$ \\
        Gemini-3.1-Pro-Preview          & \checkmark & $\times$   & $\mathbf{82.47}$ & $\mathbf{89.69}$ \\
        \midrule
        Qwen3VL-4B-Thinking             & $\times$   & $\times$   & $52.15$ & $56.68$ \\
        Qwen3VL-8B-Instruct             & $\times$   & $\times$   & $53.60$ & $58.83$ \\
        Qwen3.5-9B                      & $\times$   & $\times$   & $61.57$ & $67.69$ \\
        Qwen3VL-8B-Instruct (FT)        & $\times$   & \checkmark & $64.92$ & $71.77$ \\
        Qwen3VL-4B-Thinking (SFT + RL)  & $\times$   & \checkmark & $\mathbf{72.76}$ & $\mathbf{81.06}$ \\
        \bottomrule
    \end{tabular}}
    \label{tab:si_results}
    \vspace{-2mm}
\end{table}

\begin{figure}[t]
    \centering
    \includegraphics[width=\linewidth]{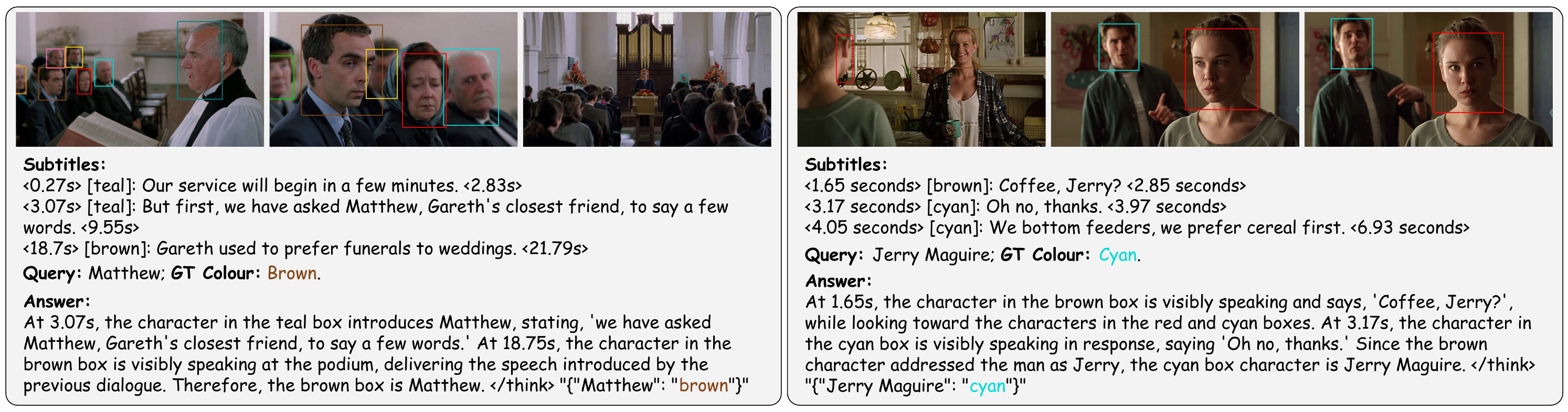}
    \vspace{-6mm}
       \caption{\textbf{Qualitative results for Character Naming.} Examples are from \textit{Four Weddings and a Funeral} and \textit{Jerry Maguire}. In each case, our default ``Qwen3VL-4B-Thinking (SFT + RL)'' model reasons jointly over subtitles and video frames to link a queried name to the correct face --- in the first, by following a verbal introduction to a later podium shot; in the second, by resolving direct address across a shot--reverse-shot exchange.}
    \label{fig:cn_visualization}
\end{figure}

\begin{table}[t]
    \centering
    \caption{\textbf{Character naming accuracy on the MCN test subset from unseen movies.} We report the accuracy results on a test subset derived from unseen movies (released after the models were trained) by all evaluated models to isolate the effects of knowledge leakage. Overall accuracy includes \texttt{none} labels, and known accuracy is restricted to queries with a valid face track.}
    \vspace{-1mm}
    \setlength{\tabcolsep}{6pt}
    \resizebox{0.83\linewidth}{!}{%
    \begin{tabular}{lcccc}
        \toprule
        \tabtitle{Model} & \tabtitle{Propriet.} & \tabtitle{Fine-Tuned} & \tabtitle{Overall Acc.} & \tabtitle{Known Acc.} \\
        \midrule
        Random Baseline                 & $\times$ & $\times$   & $18.42$ & $20.93$ \\
        \midrule
        Gemini-3-Flash-Preview          & \checkmark & $\times$   & $78.62$ & $\mathbf{81.34}$ \\
        Gemini-3.1-Pro-Preview          & \checkmark & $\times$   & $\mathbf{80.50}$ & $78.36$ \\
        \midrule
        Qwen3VL-8B-Instruct             & $\times$   & $\times$   & $45.28$ & $45.52$ \\
        Qwen3.5-9B                      & $\times$   & $\times$   & $50.94$ & $54.48$ \\
        Qwen3VL-4B-Thinking (SFT + RL)  & $\times$   & \checkmark & $\mathbf{64.78}$ & $\mathbf{71.64}$ \\
        \bottomrule
    \end{tabular}}
    \label{tab:2026_results}
    \vspace{-2mm}
\end{table}

\textbf{Results.}
We report Character Naming results in Tab.~\ref{tab:si_results}. The pretrained Qwen3VL-8B-Instruct achieves $53.60\%$ overall accuracy, improving to $64.92\%$ via naive fine-tuning. In contrast, our two-stage SFT+GRPO approach with \emph{color-permutation augmentation}, applied to a smaller 4B-Thinking backbone, yields a $+7.84\%$ gain over the naive 8B baseline. Despite the smaller model size, our approach closes most of the gap to Gemini-3.1-Pro-Preview and clearly outperforms all open-source alternatives. Qualitative examples are shown in Fig.~\ref{fig:cn_visualization}.

\textbf{Discussion.}
Compared with open-source models such as Qwen3.5-9B, our two-stage fine-tuned model more accurately names characters in direct address scenarios (e.g.\ shot--reverse-shot, telephone conversations), better exploits visual cues (e.g.\ pointing, handshakes), and improves temporal grounding between subtitles and frames. A qualitative comparison is provided in Appendix~\ref{supp:cn_example}.

Nevertheless, a gap remains to the strongest proprietary model, likely driven by errors in transcription and active-speaker detection. We hypothesise that models such as Gemini~\cite{gemini3} can partially recover from these errors by leveraging lip motion, whereas our model is more sensitive to such upstream noise.

There are also cases where both Gemini~\cite{gemini3} and our model fail. In particular, both struggle with (i) subtle cues (e.g.\ brief pointing among multiple characters), (ii) complex scenes with motion, and (iii) over-predicting on-screen characters when the ground-truth character is absent. We refer readers to Appendix~\ref{supp:cn_example} for further analysis of these failure modes.

\textbf{Evaluation on Unseen Movies.} We also observe that proprietary models and pre-trained open-source models might have seen the movies that we test on. Therefore, they could infer the character identities without using cues from the clips but from their previous knowledge. This introduces a risk of data contamination, potentially leading to an overestimation of the models' actual capabilities on our benchmark. To address this concern, we collect $102$ manually annotated clips from four films released from March 2026 onward (\textit{Voicemails for Isabelle}, 19 June 2026; \textit{In the Grey}, 15 May 2026; \textit{The Death of Robin Hood}, 19 June 2026; \textit{Heartstopper Forever}, 17 July 2026), after the reported training cut-offs of both Gemini-3.1-Pro (February 2026) and Qwen3-VL (September 2025), yielding $159$ character-naming questions, $25$ of which have none as the answer. Neither model can therefore have encountered these films during pre-training.

In Tab.~\ref{tab:2026_results}, the relative ordering of methods from Tab.~\ref{tab:si_results} is unchanged on both metrics. Our fine-tuned 4B model again outperforms all open-source baselines by a wide margin and trails only the proprietary Gemini models. Accuracy drops for all methods by a margin comparable to that of the random baseline, reflecting the greater difficulty of these clips, which frequently feature fast-moving action and dense background crowds, compared to the MCN test set used in Tab.~\ref{tab:si_results}. Our model retains most of its original accuracy, confirming that it relies on in-clip reasoning rather than prior familiarity with the films.

\subsection{Results on Streaming Character Bank Construction}
\label{exp_linkage}

\begin{table}[t]
    \centering
    \caption{\textbf{Results on the MCN First Occurrence dataset.} We report mean Temporal IoU (mTIoU) against two reference intervals, first visual occurrence and first naming, and Principal Character Recall (PCR@$K$) on the final character bank for $K\in\{5,10,15\}$.
    Our method operates in a streaming setting, whereas proprietary baselines are given full-movie access.}
    \vspace{-1mm}
    \setlength{\tabcolsep}{6pt}
    \resizebox{0.93\linewidth}{!}{%
    \begin{tabular}{lcccccc}
        \toprule
        \tabtitle{Method} & \tabtitle{Streaming} & \tabtitle{mTIoU$_v$} & \tabtitle{mTIoU$_n$} & \tabtitle{PCR@5} & \tabtitle{PCR@10}  & \tabtitle{PCR@15} \\
        \midrule
        Gemini-3-Flash-Preview   & $\times$   & $51.81$ & $30.92$ & $35.00$ & $32.04$ & $29.84$ \\
        Gemini-3.1-Pro-Preview   & $\times$   & $49.05$ & $30.10$ & $44.00$ & $37.15$ & $34.93$\\
        \midrule
        \textbf{Ours} & \checkmark & $\mathbf{84.90}$ & $\mathbf{54.40}$ & $\mathbf{63.00}$ & $\mathbf{54.54}$ & $\mathbf{48.19}$ \\
        \bottomrule
    \end{tabular}}
    \vspace{-2mm}
    \label{tab:linkage_results}
\end{table}

\paragraph{Baseline.} For streaming character bank construction, we use Gemini-3-Flash-Preview and Gemini-3.1-Pro-Preview~\cite{gemini3} as baselines, prompting them to output each character’s name, first visual occurrence timestamp, and first naming timestamp. To obtain face clusters, we ask the models to identify timestamps with close-up or clear solo shots, from which we perform face detection, extract features, and cluster faces, selecting the largest connected component as the predicted cluster. As these models are not designed to directly output face crops or clusters, we consider a prediction correct if any face in the cluster matches the ground-truth identity.

\textbf{Results.}
As shown in Tab.~\ref{tab:linkage_results}, our pipeline substantially outperforms both proprietary baselines across all metrics, despite operating in the more challenging streaming setting. These improvements reflect more accurate identification of when characters first appear and are named (mTIoU), as well as stronger alignment of the final character bank with canonical identities from external sources (PCR). These gains are especially notable given that the baselines have full-movie access while ours processes the video in a single streaming pass. Appendix~\ref{supp:qual_scbc} shows two examples of the final constructed character bank.

\textbf{Discussion.}
Our pipeline performs well on movies such as \textit{Juno} and \textit{Meet the Parents}, which feature relatively simple scenes with few participants and a limited set of recurring characters. In these cases, identity clustering is generally reliable, aided by our face-track filtering. Even when Character Naming errors occur, our method can often recover as additional evidence emerges over time.

Performance degrades, however, in movies with many scenes containing numerous minor or background characters. In \textit{Harry Potter and the Goblet of Fire}, for example, background characters introduce noise into clustering, which distracts the Character Naming process. Movies with very large named casts, such as \textit{Four Weddings and a Funeral}, with over $50$ named characters among hundreds of recurring faces, are also particularly challenging, as the model must maintain and disambiguate many identities over long temporal contexts.

\section{Related Work}
\label{related_work}

\textbf{Character-Centric Movie Understanding.}
Movie understanding has emerged as a milestone task for video understanding, owing to its long temporal span and rich multimodal narrative structure, motivating a variety of datasets \citep{huang2020movienet, tapaswi2016movieqa, bain2020condensed, soldan2022mad, vicol2018moviegraphs}. A representative line of research is \emph{Audio Description (AD) generation} \citep{rohrbach2015dataset, han2023autoad, han2023autoadii, han2024autoadiii, xie2024autoad, xie2025shot, khandelwal2025more, gui2025character}, where \citet{han2023autoadii} identifies character recognition as central to the task. Existing work typically takes one of two shortcuts: relying on curated sources like IMDB cast lists \citep{pini2019m, han2023autoad, han2023autoadii, han2024autoadiii, xie2024autoad, xie2025shot} or fan wikis \citep{gui2025character}, which are often incomplete or unavailable for new films, or sidestepping naming entirely with placeholders like ``Person~A'' \citep{fu2025video, zhou2025mlvu, song2024moviechat}. A separate line of work links names to faces directly from video. Early approaches align shooting scripts with subtitles for weak labels and use active-speaker detection to associate them with face tracks~\citep{everingham2006hello, sivic2009you}. Extensions add probabilistic models over face, speech, and clothing cues~\citep{tapaswi2012knock}, coreference resolution~\citep{ramanathan2014linking}, and script-free dialogue-based naming for recurring TV casts~\citep{haurilet2016naming, nagrani2018benedict, cour2010talking}. A complementary line assumes identities are known and analyses interactions via gaze and eyelines~\citep{marin2014detecting, marin2019laeo, gu2018ava, recasens2017following}. We revisit this problem in the era of MLLMs. Prior dialogue-based approaches, including \citet{haurilet2016naming}, cast naming as an optimisation over many mentions across clips or episodes of a recurring TV cast; we instead resolve identity from the single clip surrounding the name as a human viewer does, generalising to feature-length movies in the wild where names are typically mined from one-off dialogue. In concurrent work, \citet{pu2026omniscript} track identities via textual appearance descriptions, a lossy representation that fails to distinguish visually similar characters and degrades under costume or lighting changes. Our character bank instead grounds each identity in visual exemplars, retaining the fine-grained cues text abstracts away.

\textbf{Visual Reasoning in Vision-Language Models (MLLMs).}
Recent work extends reasoning from Large Language Models (LLMs) \citep{wei2022chain, kojima2022large} to the multimodal setting. Early approaches prompt frozen MLLMs to produce intermediate reasoning steps \citep{wang2025multimodal, mitra2024compositional}, while more recent methods train explicit reasoning via reinforcement learning. Inspired by \citep{guo2025deepseek}, Group Relative Policy Optimization (GRPO) has been adapted to MLLMs, showing that RL with verifiable rewards improves reasoning ability \citep{shen2025vlm, huang2025vision, zhou2025r1, meng2025mm}, including across perception tasks such as detection
\citep{liu2025visual}, segmentation \citep{liu2025seg}, and fine-grained perception \citep{yu2025perception}. For video understanding specifically, prior work incorporates temporal reasoning into this paradigm, e.g. via temporal-aware GRPO \citep{feng2025video}, joint spatio-temporal reasoning \citep{li2025videochat}, and long-form temporal grounding \citep{wang2025time}. In parallel, visual prompting introduces structured cues, such as colored masks~\citep{yang2023set}, boxes~\citep{shtedritski2023does, cai2024vip}, region tokens~\citep{chen2023shikra, you2023ferret}, or identity markers~\citep{xie2024autoad}, to link visual entities with language. Our approach builds on both lines of work, combining color-based visual prompting with a two-stage SFT + GRPO fine-tuning strategy to enable an MLLM to reason about character identity from visual and dialogue cues jointly.

\textbf{Person Clustering in Long Videos.}
Grouping person appearances by identity is a long-standing problem in video understanding, with applications to cast listing, character-based browsing, and story understanding \citep{fitzgibbon2002affine, cinbis2011unsupervised, everingham2006hello, tapaswi2019video}.
Early work focuses on \emph{face clustering}, using metric or representation learning to group face tracks under variations in pose, lighting, and expression~\citep{sharma2019self, wang2019linkage, yang2019learning, shen2021structure}. These methods exploit spatio-temporal structure in edited videos via co-occurrence constraints~\citep{bauml2013semi, kalogeiton2020constrained, wu2013simultaneous}, or apply nearest-neighbor clustering over pretrained embeddings~\citep{sarfraz2019efficient, otto2017clustering}. However, they rely on visible faces, limiting performance when characters are occluded, off-screen, or viewed from behind.
To address this, \citet{brown2021face} introduce Video Person-Clustering, grouping face, body, and voice cues via multimodal hierarchical clustering, with later work extending this direction using graph-based formulations \citep{nagrani2018learnable}.
Our pipeline follows a similar multimodal approach, using face embeddings as the primary signal with voice as fallback. Unlike prior work, we operate in a \emph{streaming} setting, where clusters are instantiated, updated, and named as the film unfolds, extending online clustering~\citep{aggarwal2003framework, cao2006density} to feature-length videos.

\section{Discussion}
\label{sec:discussion}

In this paper, we introduce the task of \emph{Character Naming} from dialogue and close the gap between open-source models and the best proprietary models. Building on this task, we further introduce a streaming pipeline to construct a \emph{character bank}, enabling online, human-like movie understanding. For both tasks, we compare our method with Gemini~\cite{gemini3}, though neither task is fully solved, even by proprietary models. A natural next step is to develop models or an agentic framework to overcome the current failure modes, and thereby improve performance.

\subsubsection*{Acknowledgements}
This research is supported by UK EPSRC Programme Grant VisualAI (EP/T028572/1), a Royal Society Research Professorship RSRP\textbackslash R\textbackslash 241003, and Clarendon Scholarships.

{
\small
\bibliographystyle{plainnat}
\bibliography{references}

@article{carion2025sam,
  title={Sam 3: Segment anything with concepts},
  author={Carion, Nicolas and Gustafson, Laura and Hu, Yuan-Ting and Debnath, Shoubhik and Hu, Ronghang and Suris, Didac and Ryali, Chaitanya and Alwala, Kalyan Vasudev and Khedr, Haitham and Huang, Andrew and others},
  journal={arXiv preprint arXiv:2511.16719},
  year={2025}
}

@inproceedings{Deng2020CVPR,
title = {RetinaFace: Single-Shot Multi-Level Face Localisation in the Wild},
author = {Deng, Jiankang and Guo, Jia and Ververas, Evangelos and Kotsia, Irene and Zafeiriou, Stefanos},
booktitle = {CVPR},
year = {2020}
}

@inproceedings{brown2021face,
  title={Face, body, voice: Video person-clustering with multiple modalities},
  author={Brown, Andrew and Kalogeiton, Vicky and Zisserman, Andrew},
  booktitle={proceedings of the IEEE/CVF International Conference on Computer Vision},
  pages={3184--3194},
  year={2021}
}

@misc{pyscenedetect,
  author       = {Castellano, Brandon},
  title        = {{PySceneDetect}: Video Scene Cut Detection and Analysis Tool},
  year         = {2024},
  howpublished = {\url{https://github.com/Breakthrough/PySceneDetect}},
  note         = {Accessed: 2026-04-19}
}

@inproceedings{rekesh2023fast,
  title={Fast conformer with linearly scalable attention for efficient speech recognition},
  author={Rekesh, Dima and Koluguri, Nithin Rao and Kriman, Samuel and Majumdar, Somshubra and Noroozi, Vahid and Huang, He and Hrinchuk, Oleksii and Puvvada, Krishna and Kumar, Ankur and Balam, Jagadeesh and others},
  booktitle={2023 IEEE Automatic Speech Recognition and Understanding Workshop (ASRU)},
  pages={1--8},
  year={2023},
  organization={IEEE}
}

@inproceedings{tao2021someone,
  title={Is someone speaking? exploring long-term temporal features for audio-visual active speaker detection},
  author={Tao, Ruijie and Pan, Zexu and Das, Rohan Kumar and Qian, Xinyuan and Shou, Mike Zheng and Li, Haizhou},
  booktitle={Proceedings of the 29th ACM international conference on multimedia},
  pages={3927--3935},
  year={2021}
}

@article{han2025efficient,
  title={Efficient and Generalizable Speaker Diarization via Structured Pruning of Self-Supervised Models},
  author={Han, Jiangyu and P{\'a}lka, Petr and Delcroix, Marc and Landini, Federico and Rohdin, Johan and Cernock{\`y}, Jan and Burget, Luk{\'a}{\v{s}}},
  journal={arXiv preprint arXiv:2506.18623},
  year={2025}
}

@inproceedings{xie2024autoad,
  title={Autoad-zero: A training-free framework for zero-shot audio description},
  author={Xie, Junyu and Han, Tengda and Bain, Max and Nagrani, Arsha and Varol, G{\"u}l and Xie, Weidi and Zisserman, Andrew},
  booktitle={Proceedings of the Asian Conference on Computer Vision},
  pages={2265--2281},
  year={2024}
}

@article{bai2025qwen3,
  title={Qwen3-vl technical report},
  author={Bai, Shuai and Cai, Yuxuan and Chen, Ruizhe and Chen, Keqin and Chen, Xionghui and Cheng, Zesen and Deng, Lianghao and Ding, Wei and Gao, Chang and Ge, Chunjiang and others},
  journal={arXiv preprint arXiv:2511.21631},
  year={2025}
}

@article{guo2025deepseek,
  title={Deepseek-r1: Incentivizing reasoning capability in llms via reinforcement learning},
  author={Guo, Daya and Yang, Dejian and Zhang, Haowei and Song, Junxiao and Wang, Peiyi and Zhu, Qihao and Xu, Runxin and Zhang, Ruoyu and Ma, Shirong and Bi, Xiao and others},
  journal={arXiv preprint arXiv:2501.12948},
  year={2025}
}

@inproceedings{huang2020movienet,
  title={Movienet: A holistic dataset for movie understanding},
  author={Huang, Qingqiu and Xiong, Yu and Rao, Anyi and Wang, Jiaze and Lin, Dahua},
  booktitle={European conference on computer vision},
  pages={709--727},
  year={2020},
  organization={Springer}
}

@techreport{gemini3,
  author      = {{Google DeepMind}},
  title       = {Gemini 3 {Pro} Model Card},
  institution = {Google},
  year        = {2025},
  month       = {December},
  url         = {https://storage.googleapis.com/deepmind-media/Model-Cards/Gemini-3-Pro-Model-Card.pdf}
}

@article{desplanques2020ecapa,
  title={Ecapa-tdnn: Emphasized channel attention, propagation and aggregation in tdnn based speaker verification},
  author={Desplanques, Brecht and Thienpondt, Jenthe and Demuynck, Kris},
  journal={arXiv preprint arXiv:2005.07143},
  year={2020}
}

@inproceedings{han2023autoad,
  title={Autoad: Movie description in context},
  author={Han, Tengda and Bain, Max and Nagrani, Arsha and Varol, G{\"u}l and Xie, Weidi and Zisserman, Andrew},
  booktitle={Proceedings of the IEEE/CVF Conference on Computer Vision and Pattern Recognition},
  pages={18930--18940},
  year={2023}
}

@inproceedings{han2023autoadii,
  title={Autoad ii: The sequel-who, when, and what in movie audio description},
  author={Han, Tengda and Bain, Max and Nagrani, Arsha and Varol, Gul and Xie, Weidi and Zisserman, Andrew},
  booktitle={Proceedings of the IEEE/CVF International Conference on Computer Vision},
  pages={13645--13655},
  year={2023}
}

@inproceedings{han2024autoadiii,
  title={Autoad iii: The prequel-back to the pixels},
  author={Han, Tengda and Bain, Max and Nagrani, Arsha and Varol, G{\"u}l and Xie, Weidi and Zisserman, Andrew},
  booktitle={Proceedings of the IEEE/CVF Conference on Computer Vision and Pattern Recognition},
  pages={18164--18174},
  year={2024}
}

@inproceedings{xie2025shot,
  title={Shot-by-Shot: Film-Grammar-Aware Training-Free Audio Description Generation},
  author={Xie, Junyu and Han, Tengda and Bain, Max and Nagrani, Arsha and Khandelwal, Eshika and Varol, G{\"u}l and Xie, Weidi and Zisserman, Andrew},
  booktitle={Proceedings of the IEEE/CVF International Conference on Computer Vision},
  pages={16503--16513},
  year={2025}
}

@inproceedings{gui2025character,
  title={Character-Centric Understanding of Animated Movies},
  author={Gui, Zhongrui and Xie, Junyu and Han, Tengda and Xie, Weidi and Zisserman, Andrew},
  booktitle={Proceedings of the 33rd ACM International Conference on Multimedia},
  pages={3300--3309},
  year={2025}
}

@article{khandelwal2025more,
  title={More than a Moment: Towards Coherent Sequences of Audio Descriptions},
  author={Khandelwal, Eshika and Xie, Junyu and Han, Tengda and Bain, Max and Nagrani, Arsha and Zisserman, Andrew and Varol, G{\"u}l and Tapaswi, Makarand},
  journal={arXiv preprint arXiv:2510.25440},
  year={2025}
}

@inproceedings{tapaswi2016movieqa,
  title={Movieqa: Understanding stories in movies through question-answering},
  author={Tapaswi, Makarand and Zhu, Yukun and Stiefelhagen, Rainer and Torralba, Antonio and Urtasun, Raquel and Fidler, Sanja},
  booktitle={Proceedings of the IEEE conference on computer vision and pattern recognition},
  pages={4631--4640},
  year={2016}
}

@inproceedings{fu2025video,
  title={Video-mme: The first-ever comprehensive evaluation benchmark of multi-modal llms in video analysis},
  author={Fu, Chaoyou and Dai, Yuhan and Luo, Yongdong and Li, Lei and Ren, Shuhuai and Zhang, Renrui and Wang, Zihan and Zhou, Chenyu and Shen, Yunhang and Zhang, Mengdan and others},
  booktitle={Proceedings of the IEEE/CVF conference on computer vision and pattern recognition},
  pages={24108--24118},
  year={2025}
}

@inproceedings{zhou2025mlvu,
  title={Mlvu: Benchmarking multi-task long video understanding},
  author={Zhou, Junjie and Shu, Yan and Zhao, Bo and Wu, Boya and Liang, Zhengyang and Xiao, Shitao and Qin, Minghao and Yang, Xi and Xiong, Yongping and Zhang, Bo and others},
  booktitle={Proceedings of the IEEE/CVF Conference on Computer Vision and Pattern Recognition},
  pages={13691--13701},
  year={2025}
}

@inproceedings{song2024moviechat,
  title={Moviechat: From dense token to sparse memory for long video understanding},
  author={Song, Enxin and Chai, Wenhao and Wang, Guanhong and Zhang, Yucheng and Zhou, Haoyang and Wu, Feiyang and Chi, Haozhe and Guo, Xun and Ye, Tian and Zhang, Yanting and others},
  booktitle={Proceedings of the IEEE/CVF Conference on Computer Vision and Pattern Recognition},
  pages={18221--18232},
  year={2024}
}

@article{wei2022chain,
  title={Chain-of-thought prompting elicits reasoning in large language models},
  author={Wei, Jason and Wang, Xuezhi and Schuurmans, Dale and Bosma, Maarten and Xia, Fei and Chi, Ed and Le, Quoc V and Zhou, Denny and others},
  journal={Advances in neural information processing systems},
  volume={35},
  pages={24824--24837},
  year={2022}
}

@article{kojima2022large,
  title={Large language models are zero-shot reasoners},
  author={Kojima, Takeshi and Gu, Shixiang Shane and Reid, Machel and Matsuo, Yutaka and Iwasawa, Yusuke},
  journal={Advances in neural information processing systems},
  volume={35},
  pages={22199--22213},
  year={2022}
}

@article{wang2025multimodal,
  title={Multimodal chain-of-thought reasoning: A comprehensive survey},
  author={Wang, Yaoting and Wu, Shengqiong and Zhang, Yuecheng and Yan, Shuicheng and Liu, Ziwei and Luo, Jiebo and Fei, Hao},
  journal={arXiv preprint arXiv:2503.12605},
  year={2025}
}

@inproceedings{mitra2024compositional,
  title={Compositional chain-of-thought prompting for large multimodal models},
  author={Mitra, Chancharik and Huang, Brandon and Darrell, Trevor and Herzig, Roei},
  booktitle={Proceedings of the IEEE/CVF Conference on Computer Vision and Pattern Recognition},
  pages={14420--14431},
  year={2024}
}

@article{shen2025vlm,
  title={Vlm-r1: A stable and generalizable r1-style large vision-language model},
  author={Shen, Haozhan and Liu, Peng and Li, Jingcheng and Fang, Chunxin and Ma, Yibo and Liao, Jiajia and Shen, Qiaoli and Zhang, Zilun and Zhao, Kangjia and Zhang, Qianqian and others},
  journal={arXiv preprint arXiv:2504.07615},
  year={2025}
}

@article{huang2025vision,
  title={Vision-r1: Incentivizing reasoning capability in multimodal large language models},
  author={Huang, Wenxuan and Jia, Bohan and Zhai, Zijie and Cao, Shaosheng and Ye, Zheyu and Zhao, Fei and Xu, Zhe and Tang, Xu and Hu, Yao and Lin, Shaohui},
  journal={arXiv preprint arXiv:2503.06749},
  year={2025}
}

@article{feng2025video,
  title={Video-r1: Reinforcing video reasoning in mllms},
  author={Feng, Kaituo and Gong, Kaixiong and Li, Bohao and Guo, Zonghao and Wang, Yibing and Peng, Tianshuo and Wu, Junfei and Zhang, Xiaoying and Wang, Benyou and Yue, Xiangyu},
  journal={arXiv preprint arXiv:2503.21776},
  year={2025}
}

@article{li2025videochat,
  title={Videochat-r1: Enhancing spatio-temporal perception via reinforcement fine-tuning},
  author={Li, Xinhao and Yan, Ziang and Meng, Desen and Dong, Lu and Zeng, Xiangyu and He, Yinan and Wang, Yali and Qiao, Yu and Wang, Yi and Wang, Limin},
  journal={arXiv preprint arXiv:2504.06958},
  year={2025}
}

@article{yang2023set,
  title={Set-of-mark prompting unleashes extraordinary visual grounding in gpt-4v},
  author={Yang, Jianwei and Zhang, Hao and Li, Feng and Zou, Xueyan and Li, Chunyuan and Gao, Jianfeng},
  journal={arXiv preprint arXiv:2310.11441},
  year={2023}
}

@inproceedings{shtedritski2023does,
  title={What does clip know about a red circle? visual prompt engineering for vlms},
  author={Shtedritski, Aleksandar and Rupprecht, Christian and Vedaldi, Andrea},
  booktitle={Proceedings of the IEEE/CVF International Conference on Computer Vision},
  pages={11987--11997},
  year={2023}
}

@inproceedings{fitzgibbon2002affine,
  title={On affine invariant clustering and automatic cast listing in movies},
  author={Fitzgibbon, Andrew and Zisserman, Andrew},
  booktitle={European Conference on Computer Vision},
  pages={304--320},
  year={2002},
  organization={Springer}
}

@inproceedings{cinbis2011unsupervised,
  title={Unsupervised metric learning for face identification in TV video},
  author={Cinbis, Ramazan Gokberk and Verbeek, Jakob and Schmid, Cordelia},
  booktitle={2011 International Conference on Computer Vision},
  pages={1559--1566},
  year={2011},
  organization={IEEE}
}

@inproceedings{tapaswi2019video,
  title={Video face clustering with unknown number of clusters},
  author={Tapaswi, Makarand and Law, Marc T and Fidler, Sanja},
  booktitle={Proceedings of the IEEE/CVF International Conference on Computer Vision},
  pages={5027--5036},
  year={2019}
}

@inproceedings{sharma2019self,
  title={Self-supervised learning of face representations for video face clustering},
  author={Sharma, Vivek and Tapaswi, Makarand and Sarfraz, M Saquib and Stiefelhagen, Rainer},
  booktitle={2019 14th IEEE International Conference on Automatic Face \& Gesture Recognition (FG 2019)},
  pages={1--8},
  year={2019},
  organization={IEEE}
}

@inproceedings{bauml2013semi,
  title={Semi-supervised learning with constraints for person identification in multimedia data},
  author={Bauml, Martin and Tapaswi, Makarand and Stiefelhagen, Rainer},
  booktitle={Proceedings of the IEEE conference on computer vision and pattern recognition},
  pages={3602--3609},
  year={2013}
}

@article{kalogeiton2020constrained,
  title={Constrained video face clustering using 1nn relations},
  author={Kalogeiton, Vicky and Zisserman, Andrew},
  journal={https://ora.ox.ac.uk/objects/uuid:23f369b3-794b-42a6-a6af-ab4b0d726ba4/files/spn89d688d},
  year={2020}
}

@inproceedings{sarfraz2019efficient,
  title={Efficient parameter-free clustering using first neighbor relations},
  author={Sarfraz, Saquib and Sharma, Vivek and Stiefelhagen, Rainer},
  booktitle={Proceedings of the IEEE/CVF conference on computer vision and pattern recognition},
  pages={8934--8943},
  year={2019}
}

@inproceedings{bain2020condensed,
  title={Condensed movies: Story based retrieval with contextual embeddings},
  author={Bain, Max and Nagrani, Arsha and Brown, Andrew and Zisserman, Andrew},
  booktitle={Proceedings of the Asian Conference on Computer Vision},
  year={2020}
}

@inproceedings{soldan2022mad,
  title={Mad: A scalable dataset for language grounding in videos from movie audio descriptions},
  author={Soldan, Mattia and Pardo, Alejandro and Alc{\'a}zar, Juan Le{\'o}n and Caba, Fabian and Zhao, Chen and Giancola, Silvio and Ghanem, Bernard},
  booktitle={Proceedings of the IEEE/CVF Conference on Computer Vision and Pattern Recognition},
  pages={5026--5035},
  year={2022}
}

@inproceedings{rohrbach2015dataset,
  title={A dataset for movie description},
  author={Rohrbach, Anna and Rohrbach, Marcus and Tandon, Niket and Schiele, Bernt},
  booktitle={Proceedings of the IEEE conference on computer vision and pattern recognition},
  pages={3202--3212},
  year={2015}
}

@article{pini2019m,
  title={M-VAD names: a dataset for video captioning with naming},
  author={Pini, Stefano and Cornia, Marcella and Bolelli, Federico and Baraldi, Lorenzo and Cucchiara, Rita},
  journal={Multimedia Tools and Applications},
  volume={78},
  number={10},
  pages={14007--14027},
  year={2019},
  publisher={Springer}
}

@inproceedings{sivic2009you,
  title={“Who are you?”-Learning person specific classifiers from video},
  author={Sivic, Josef and Everingham, Mark and Zisserman, Andrew},
  booktitle={2009 IEEE Conference on Computer Vision and Pattern Recognition},
  pages={1145--1152},
  year={2009},
  organization={IEEE}
}

@inproceedings{everingham2006hello,
  title={Hello! My name is... Buffy''--Automatic Naming of Characters in TV Video.},
  author={Everingham, Mark and Sivic, Josef and Zisserman, Andrew},
  booktitle={BMVC},
  volume={2},
  number={4},
  pages={6},
  year={2006}
}

@inproceedings{ramanathan2014linking,
  title={Linking people in videos with “their” names using coreference resolution},
  author={Ramanathan, Vignesh and Joulin, Armand and Liang, Percy and Fei-Fei, Li},
  booktitle={European conference on computer vision},
  pages={95--110},
  year={2014},
  organization={Springer}
}

@inproceedings{tapaswi2012knock,
  title={“Knock! Knock! Who is it?” probabilistic person identification in TV-series},
  author={Tapaswi, Makarand and B{\"a}uml, Martin and Stiefelhagen, Rainer},
  booktitle={2012 IEEE Conference on Computer Vision and Pattern Recognition},
  pages={2658--2665},
  year={2012},
  organization={IEEE}
}

@article{nagrani2018benedict,
  title={From benedict cumberbatch to sherlock holmes: Character identification in tv series without a script},
  author={Nagrani, Arsha and Zisserman, Andrew},
  journal={arXiv preprint arXiv:1801.10442},
  year={2018}
}

@inproceedings{haurilet2016naming,
  title={Naming TV characters by watching and analyzing dialogs},
  author={Haurilet, Monica-Laura and Tapaswi, Makarand and Al-Halah, Ziad and Stiefelhagen, Rainer},
  booktitle={2016 IEEE winter conference on applications of computer vision (WACV)},
  pages={1--9},
  year={2016},
  organization={IEEE}
}

@inproceedings{vicol2018moviegraphs,
  title={Moviegraphs: Towards understanding human-centric situations from videos},
  author={Vicol, Paul and Tapaswi, Makarand and Castrejon, Lluis and Fidler, Sanja},
  booktitle={Proceedings of the IEEE conference on computer vision and pattern recognition},
  pages={8581--8590},
  year={2018}
}

@article{zhou2025r1,
  title={R1-Zero's" Aha Moment" in Visual Reasoning on a 2B Non-SFT Model},
  author={Zhou, Hengguang and Li, Xirui and Wang, Ruochen and Cheng, Minhao and Zhou, Tianyi and Hsieh, Cho-Jui},
  journal={arXiv preprint arXiv:2503.05132},
  year={2025}
}

@article{meng2025mm,
  title={Mm-eureka: Exploring the frontiers of multimodal reasoning with rule-based reinforcement learning},
  author={Meng, Fanqing and Du, Lingxiao and Liu, Zongkai and Zhou, Zhixiang and Lu, Quanfeng and Fu, Daocheng and Han, Tiancheng and Shi, Botian and Wang, Wenhai and He, Junjun and others},
  journal={arXiv preprint arXiv:2503.07365},
  year={2025}
}

@inproceedings{liu2025visual,
  title={Visual-rft: Visual reinforcement fine-tuning},
  author={Liu, Ziyu and Sun, Zeyi and Zang, Yuhang and Dong, Xiaoyi and Cao, Yuhang and Duan, Haodong and Lin, Dahua and Wang, Jiaqi},
  booktitle={Proceedings of the IEEE/CVF International Conference on Computer Vision},
  pages={2034--2044},
  year={2025}
}

@article{liu2025seg,
  title={Seg-zero: Reasoning-chain guided segmentation via cognitive reinforcement},
  author={Liu, Yuqi and Peng, Bohao and Zhong, Zhisheng and Yue, Zihao and Lu, Fanbin and Yu, Bei and Jia, Jiaya},
  journal={arXiv preprint arXiv:2503.06520},
  year={2025}
}

@article{yu2025perception,
  title={Perception-r1: Pioneering perception policy with reinforcement learning},
  author={Yu, En and Lin, Kangheng and Zhao, Liang and Yin, Jisheng and Wei, Yana and Peng, Yuang and Wei, Haoran and Sun, Jianjian and Han, Chunrui and Ge, Zheng and others},
  journal={arXiv preprint arXiv:2504.07954},
  year={2025}
}

@article{wang2025time,
  title={Time-r1: Post-training large vision language model for temporal video grounding},
  author={Wang, Ye and Wang, Ziheng and Xu, Boshen and Du, Yang and Lin, Kejun and Xiao, Zihan and Yue, Zihao and Ju, Jianzhong and Zhang, Liang and Yang, Dingyi and others},
  journal={arXiv preprint arXiv:2503.13377},
  year={2025}
}

@inproceedings{cai2024vip,
  title={Vip-llava: Making large multimodal models understand arbitrary visual prompts},
  author={Cai, Mu and Liu, Haotian and Mustikovela, Siva Karthik and Meyer, Gregory P and Chai, Yuning and Park, Dennis and Lee, Yong Jae},
  booktitle={Proceedings of the IEEE/CVF Conference on Computer Vision and Pattern Recognition},
  pages={12914--12923},
  year={2024}
}

@article{chen2023shikra,
  title={Shikra: Unleashing multimodal llm's referential dialogue magic},
  author={Chen, Keqin and Zhang, Zhao and Zeng, Weili and Zhang, Richong and Zhu, Feng and Zhao, Rui},
  journal={arXiv preprint arXiv:2306.15195},
  year={2023}
}

@article{you2023ferret,
  title={Ferret: Refer and ground anything anywhere at any granularity},
  author={You, Haoxuan and Zhang, Haotian and Gan, Zhe and Du, Xianzhi and Zhang, Bowen and Wang, Zirui and Cao, Liangliang and Chang, Shih-Fu and Yang, Yinfei},
  journal={arXiv preprint arXiv:2310.07704},
  year={2023}
}

@inproceedings{wang2019linkage,
  title={Linkage based face clustering via graph convolution network},
  author={Wang, Zhongdao and Zheng, Liang and Li, Yali and Wang, Shengjin},
  booktitle={Proceedings of the IEEE/CVF conference on computer vision and pattern recognition},
  pages={1117--1125},
  year={2019}
}

@inproceedings{yang2019learning,
  title={Learning to cluster faces on an affinity graph},
  author={Yang, Lei and Zhan, Xiaohang and Chen, Dapeng and Yan, Junjie and Loy, Chen Change and Lin, Dahua},
  booktitle={Proceedings of the IEEE/CVF conference on computer vision and pattern recognition},
  pages={2298--2306},
  year={2019}
}

@inproceedings{shen2021structure,
  title={Structure-aware face clustering on a large-scale graph with 107 nodes},
  author={Shen, Shuai and Li, Wanhua and Zhu, Zheng and Huang, Guan and Du, Dalong and Lu, Jiwen and Zhou, Jie},
  booktitle={Proceedings of the IEEE/CVF conference on computer vision and pattern recognition},
  pages={9085--9094},
  year={2021}
}

@inproceedings{wu2013simultaneous,
  title={Simultaneous clustering and tracklet linking for multi-face tracking in videos},
  author={Wu, Baoyuan and Lyu, Siwei and Hu, Bao-Gang and Ji, Qiang},
  booktitle={Proceedings of the IEEE international conference on computer vision},
  pages={2856--2863},
  year={2013}
}

@article{otto2017clustering,
  title={Clustering millions of faces by identity},
  author={Otto, Charles and Wang, Dayong and Jain, Anil K},
  journal={IEEE transactions on pattern analysis and machine intelligence},
  volume={40},
  number={2},
  pages={289--303},
  year={2017},
  publisher={IEEE}
}

@inproceedings{nagrani2018learnable,
  title={Learnable pins: Cross-modal embeddings for person identity},
  author={Nagrani, Arsha and Albanie, Samuel and Zisserman, Andrew},
  booktitle={Proceedings of the European conference on computer vision (ECCV)},
  pages={71--88},
  year={2018}
}

@inproceedings{aggarwal2003framework,
  title={A framework for clustering evolving data streams},
  author={Aggarwal, Charu C and Philip, S Yu and Han, Jiawei and Wang, Jianyong},
  booktitle={Proceedings 2003 VLDB conference},
  pages={81--92},
  year={2003},
  organization={Elsevier}
}

@inproceedings{cao2006density,
  title={Density-based clustering over an evolving data stream with noise},
  author={Cao, Feng and Estert, Martin and Qian, Weining and Zhou, Aoying},
  booktitle={Proceedings of the 2006 SIAM international conference on data mining},
  pages={328--339},
  year={2006},
  organization={SIAM}
}

@inproceedings{yue2023movie101,
  title={Movie101: A new movie understanding benchmark},
  author={Yue, Zihao and Zhang, Qi and Hu, Anwen and Zhang, Liang and Wang, Ziheng and Jin, Qin},
  booktitle={Proceedings of the 61st Annual Meeting of the Association for Computational Linguistics (Volume 1: Long Papers)},
  pages={4669--4684},
  year={2023}
}

@inproceedings{cour2010talking,
  title={Talking pictures: Temporal grouping and dialog-supervised person recognition},
  author={Cour, Timothee and Sapp, Benjamin and Nagle, Akash and Taskar, Ben},
  booktitle={2010 IEEE Computer Society Conference on Computer Vision and Pattern Recognition},
  pages={1014--1021},
  year={2010},
  organization={IEEE}
}

@article{marin2014detecting,
  title={Detecting people looking at each other in videos},
  author={Marin-Jimenez, Manuel Jes{\'u}s and Zisserman, Andrew and Eichner, Marcin and Ferrari, Vittorio},
  journal={International Journal of Computer Vision},
  volume={106},
  number={3},
  pages={282--296},
  year={2014},
  publisher={Springer}
}

@inproceedings{marin2019laeo,
  title={Laeo-net: revisiting people looking at each other in videos},
  author={Marin-Jimenez, Manuel J and Kalogeiton, Vicky and Medina-Suarez, Pablo and Zisserman, Andrew},
  booktitle={Proceedings of the IEEE/CVF conference on computer vision and pattern recognition},
  pages={3477--3485},
  year={2019}
}

@inproceedings{gu2018ava,
  title={Ava: A video dataset of spatio-temporally localized atomic visual actions},
  author={Gu, Chunhui and Sun, Chen and Ross, David A and Vondrick, Carl and Pantofaru, Caroline and Li, Yeqing and Vijayanarasimhan, Sudheendra and Toderici, George and Ricco, Susanna and Sukthankar, Rahul and others},
  booktitle={Proceedings of the IEEE conference on computer vision and pattern recognition},
  pages={6047--6056},
  year={2018}
}

@inproceedings{recasens2017following,
  title={Following gaze in video},
  author={Recasens, Adria and Vondrick, Carl and Khosla, Aditya and Torralba, Antonio},
  booktitle={Proceedings of the IEEE International Conference on Computer Vision},
  pages={1435--1443},
  year={2017}
}

@article{pu2026omniscript,
  title={OmniScript: Towards Audio-Visual Script Generation for Long-Form Cinematic Video},
  author={Pu, Junfu and Chen, Yuxin and Wang, Teng and Shan, Ying},
  journal={arXiv preprint arXiv:2604.11102},
  year={2026}
}
}

\newpage
\appendix

\section*{Appendix}

This Appendix contains the following sections:
\begin{itemize}[leftmargin=4mm]
    \item In Appendix~\ref{supp:preprocessing}, we elaborate on the preprocessing stage, covering face track extraction and clustering as well as the linking of voice clusters to face tracks via active-speaker synchronization.
    \item In Appendix~\ref{supp:grpo}, we present the full GRPO formulation used to fine-tune our character-naming MLLM, including the composite reward, advantage normalization, and policy update.
    \item In Appendix~\ref{supp:error_corr}, we provide detailed descriptions of the error correction mechanism during streaming character bank construction.
    \item In Appendix~\ref{supp:implementation}, we specify implementation details and hyperparameter settings for clustering, fine-tuning, and streaming character bank construction.
    \item In Appendix~\ref{supp:ablation}, we conduct ablation studies on the fine-tuning strategy, color-permutation augmentation, and error-correction mechanisms to verify key design choices.
    \item In Appendix~\ref{supp:mil_baseline}, we introduce an additional baseline for streaming character bank construction that natively produces face exemplars, and is therefore better matched to our output format than the MLLM baselines.
    \item In Appendix~\ref{supp:alignment}, we provide additional details on the MCN dataset collection, including the coarse-to-fine alignment pipeline used to recover MovieNet annotations at native resolution.
    \item In Appendix~\ref{supp:dataset}, we present representative dataset examples illustrating the diversity of films and naming cues covered by the MCN benchmark.
    \item In Appendices~\ref{supp:cn_example} and~\ref{supp:qual_scbc}, we provide additional qualitative results for character naming and streaming character bank construction, including comparisons with baselines and failure cases.
    \item In Appendix~\ref{supp:compute}, we describe the compute resources used for the experiments, including MLLM fine-tuning and inference.
    \item In Appendix~\ref{supp:broader_impacts}, we discuss the broader impacts of our work, including potential risks and the scope of the released artefacts.
    \item We also include a \emph{supplementary video} that demonstrates our character naming results and visualizes the streaming character bank construction process on the movie \textit{Juno}.
\end{itemize}

\section{Details on Preprocessing}
\label{supp:preprocessing}
This section provides additional details on the preprocessing stage of Sec.~\ref{preprocessing}. We first describe how face tracks are detected, extracted, and merged into face clusters (Sec.~\ref{supp:clustering}), and then explain how voice clusters are linked to these face clusters via active-speaker synchronization (Sec.~\ref{supp:subtitle}).

\subsection{Face Track Extraction and Clustering}
\label{supp:clustering}
To obtain complete and accurate face tracks, we detect shot boundaries using SceneDetect \citep{pyscenedetect} and query SAM~3 \citep{carion2025sam} with the text prompt ``face'' at the beginning, middle, and end of each shot to handle occlusions. We then extract face features with InsightFace \citep{Deng2020CVPR} by sampling 15 frames per track. Following \citet{brown2021face}, we cluster the tracks into face clusters $\{f_i\}$ under two constraints: (1) a \emph{spatio-temporal cannot-link constraint}, preventing tracks with (partial) temporal overlap from being grouped, since co-occurring tracks in the same frame must correspond to distinct characters; and (2) a \emph{nearest-neighbor distance constraint}, which requires the distance between the features of merged tracks to fall below a threshold $\tau_{\text{cluster}}$.

\subsection{Linking Voices to Face Tracks}
\label{supp:subtitle}
We transcribe movie dialogue using Parakeet \citep{rekesh2023fast} and group voice segments into voice clusters $\{v_j\}$ via speaker diarization \citep{han2025efficient}. To match each voice cluster to a face cluster, we apply TalkNet \citep{tao2021someone} to compute a per-frame active-speaker synchronization score $s_{k, t}$, indicating how likely face track $k$ is speaking at timestamp $t$. We rely on diarization because TalkNet, which is predominantly trained on frontal faces, fails when speakers are off-screen or not facing the camera.

For each voice cluster $v_j$ with subtitle lines $\{e\}$ (each $e$ has start and end times $t^{\text{s}}_{e}$ and $t^{\text{e}}_{e}$), we aggregate the per-frame scores at the cluster level to obtain an association score $S_{i,j}$ between face cluster $f_i$ and voice cluster $v_j$:
\begin{equation}
\label{asd_score}
    s_{e,i} \;=\; \max_{\text{track } k \in f_i} \; \frac{1}{|\mathcal{T}_{k,e}|} \sum_{t \in \mathcal{T}_{k,e}} s_{k,t},
    \quad
    S_{i,j} \;=\; \max_{e \in v_j} \, s_{e,i},
    \quad
    \mathcal{T}_{k,e} = \{\, t \,:\, t^{\text{s}}_{e} \le t \le t^{\text{e}}_{e} \,\}.
\end{equation}
We assign each $v_j$ to the face cluster $f_{i^*(j)}$ with the highest score, $i^*(j) = \operatorname*{arg\,max}_i S_{i,j}$, provided that $S_{i^*(j), j}$ exceeds a threshold $\tau_{\text{ASD}}$. As an additional override, whenever $s_{k,t}$ exceeds $\tau_{\text{override}}$ for some face track $k$ within the time span of subtitle line $e$, we directly assign $e$ to the face cluster containing $k$, bypassing the diarization-based grouping. This handles cases where TalkNet provides high-confidence per-frame evidence that contradicts the diarization output.

\section{GRPO Reinforcement Learning Fine-Tuning}
\label{supp:grpo}

This section provides the full GRPO formulation summarised in Sec.~\ref{sec:finetuning}.

\textbf{Reward.} We apply GRPO \citep{guo2025deepseek} with a composite reward that combines a format reward $r_f$ and an accuracy reward $r_a$. The format reward $r_f$ is granted whenever the output $o_i$ adheres to the expected response structure. The accuracy reward $r_a$ measures how well the predicted set of name--identity pairs $\hat{\psi}(V)$ agrees with the ground-truth pairs $\psi^{*}(V)$. Letting $\hat{i}(n)$ and $i^{*}(n)$ denote the identities paired with name $n$ under $\hat{\psi}$ and $\psi^{*}$, respectively, we set
\begin{equation}
    r_a = \sum_{n \in N} \mathbf{1}\!\left[\hat{i}(n) = i^{*}(n)\right],
\end{equation}
and define the full reward as $R_i = r_f + r_a$.

\textbf{Advantage and Policy Update.} The advantage $A_i$ is computed by normalizing rewards within each group of $G$ sampled responses:
\begin{equation}
    A_i = \frac{R_i - \mathrm{mean}(\{R_j\})}{\mathrm{std}(\{R_j\})}.
\end{equation}
Following \citet{guo2025deepseek}, the policy is updated by maximising
\begin{equation}
    \mathcal{J}_{\text{GRPO}}(\theta) = \mathbb{E}_{q, \{o_i\}} \left[ \frac{1}{G} \sum_{i=1}^{G} \min\!\left( \rho_i(\theta)\, A_i,\; \mathrm{clip}\!\left(\rho_i(\theta),\, 1-\epsilon,\, 1+\epsilon\right) A_i \right) - \beta\, D_{\mathrm{KL}}\!\left(\pi_\theta \,\|\, \pi_{\text{ref}}\right) \right],
\end{equation}
where $\rho_i(\theta) = \pi_\theta(o_i \mid q) / \pi_{\theta_{\text{old}}}(o_i \mid q)$ is the importance ratio, $\epsilon$ is the clipping threshold, and $\beta$ controls the strength of the KL regularisation against the reference policy $\pi_{\text{ref}}$.

\section{Error Correction for Character Bank Construction}
\label{supp:error_corr}
\paragraph{Additional Details on Error Mitigation.}
As outlined in~\cref{sec:error_correction}, the Markovian update process $\phi_{\mathcal{B}}$ is inherently sensitive to upstream noise, where errors from components such as transcription, active-speaker detection, and identity grounding can propagate over time. To address this, we elaborate on the three interventions applied along the streaming pipeline: face-track filtering, intra-entry name merging, and inter-entry name disambiguation.

\textbf{Face-Track Filtering.}
Prior to clustering, we introduce an MLLM-based filtering stage to discard unreliable or ambiguous face tracks, thereby maintaining clean and identifiable clusters in the bank. This step removes low-quality observations, including blurred faces and extreme profile views. Empirically, embeddings derived from the retained high-quality tracks are sufficiently stable, rendering clustering errors negligible under our operating thresholds.

\textbf{Query Name Filtering.} As previously discussed, when no on-screen character actually corresponds to the candidate name, both our model and Gemini tend to over-assign one of the visible faces to an off-screen name rather than abstaining. To circumvent this, we filter the candidate name pool upstream so that off-screen names rarely reach the naming step in the first place. Concretely, we use an LLM (Qwen3.5-9B) to inspect each subtitle window and keep only those candidate names likely to refer to a character currently on screen. This requires reasoning over the flow of the surrounding dialogue rather than classifying each line in isolation, since whether a named person is actually present or merely being discussed is rarely decidable from a single utterance.

\textbf{Inter-Entry Name Disambiguation.}
Ideally, each name should be uniquely associated with a single entry; however, MLLM-induced errors can lead to conflicts across entries. We resolve these after each shot using a greedy voting procedure: for names assigned to multiple entries, the entry with the highest vote retains the name, while others revert to their next-best candidates. This process is repeated until convergence.

Overall, these mechanisms address complementary failure modes in the streaming process and prevent error accumulation as $\mathcal{B}_t$ evolves, without requiring access to past observations.

\section{Implementation Details}
\label{supp:implementation}
We set the face-clustering distance threshold to $\tau_{\text{cluster}} = 0.4$, the active-speaker association threshold to $\tau_{\text{ASD}} = -0.075$, and the per-frame override threshold to $\tau_{\text{override}} = 0.1$. We use DiariZen-Large-s80-v2 \citep{han2025efficient} as our diarization model.

For training the MLLM for speaker identity recognition, we adopt Qwen3-VL-4B-Thinking \citep{bai2025qwen3} as our base model. To construct the first-stage supervised fine-tuning (SFT) dataset, we sample approximately $10\%$ of the training set and prompt Gemini-3-Flash-Preview \citep{gemini3} to generate reasoning traces that lead to the given ground-truth name-to-track mappings. For SFT, we fine-tune only the language model component of Qwen3-VL-4B-Thinking while keeping both the vision encoder and the multimodal projector frozen, for one epoch with a batch size of $4$, a learning rate of $1 \times 10^{-5}$, and a cosine schedule with warmup ratio $0.03$. Then, we perform GRPO fine-tuning on the remaining training set for one epoch, using a learning rate of $1 \times 10^{-6}$, a batch size of $4$, $G = 4$ sampled responses per prompt, and a KL coefficient of $\beta = 0.004$.

For streaming character bank construction, we set the face matching threshold to $\tau_f = 0.55$ and the voice fallback threshold to $\tau_v = 0.25$. Voice embeddings are extracted with ECAPA-TDNN \citep{desplanques2020ecapa} over speaker-aligned subtitle segments; both face and voice centroids are maintained as L2-normalized running means over all cluster members. Candidate names are extracted from subtitles using spaCy transformer NER pipelines (\texttt{en\_core\_web\_trf}), and off-screen mentions are filtered out by Qwen3.5-9B \citep{bai2025qwen3}. The remaining names are linked to face clusters by our fine-tuned MLLM, which receives frames sampled at $2$ FPS. Low-quality faces are filtered by Qwen3-VL-8B-Instruct \citep{bai2025qwen3} with stricter prompting before they are committed to the bank.

For PCR@K evaluation, we set the face-matching threshold $\tau_{\text{eval}} = 0.55$ and use the Double Metaphone algorithm for phonetic matching between predicted canonical names and IMDB ground-truth names. The IMDB reference embedding $\bar{e}^{*}_c$ is the L2-normalized InsightFace embedding extracted from the actor's IMDB headshot.

No hyperparameter mentioned above was tuned on the test set. In the absence of a validation split, all thresholds were set following prior work: $\tau_{\text{cluster}}$, $\tau_{f}$ and $\tau_{v}$ follow the multimodal clustering steup of \citet{brown2021face}, and $\tau_{\text{ASD}}$ and $\tau_{\text{override}}$ follow TalkNet~\cite{tao2021someone}. $\tau_{\text{eval}}$ is set equal to $\tau_{f}$ (i.e.\ to $0.55$), the value conventionally used to decide whether two InsightFace embeddings correspond to the same identity.

\section{Ablation Study}
\label{supp:ablation}

We ablate the key design choices of our method on the MCN dataset to isolate their individual contributions. Using Qwen3-VL-4B-Thinking as the base model, we evaluate the fine-tuning strategy and data augmentation for character naming. We also ablate on error correction mechanisms to measure their effects on the full streaming pipeline.

\begin{table}[th]
    \centering
    \caption{\textbf{Ablation study on Character Naming.} We compare our two-stage (SFT + RL) fine-tuning of Qwen3VL-4B-Thinking against direct SFT on a larger Qwen3VL-8B-Instruct model, and further ablate the contributions of the RL stage and color-permutation augmentation.}
    \setlength{\tabcolsep}{8pt}
    \resizebox{0.85\linewidth}{!}{%
    \begin{tabular}{lccccc}
        \toprule
        \tabtitle{Model} & \tabtitle{SFT} & \tabtitle{RL} & \tabtitle{Augment.} & \tabtitle{Overall Acc.} & \tabtitle{Known Acc.} \\
        \midrule
        Qwen3VL-8B-Instruct & \checkmark & $\times$ & $\times$ & $64.92$ & $71.77$ \\
        \midrule
        Qwen3VL-4B-Thinking & \checkmark & $\times$   & $\times$ & $62.39$ & $67.98$ \\
        Qwen3VL-4B-Thinking & \checkmark & $\times$   & \checkmark & $64.16$ & $70.41$ \\
        Qwen3VL-4B-Thinking & \checkmark & \checkmark & $\times$ & $71.37$ & $79.56$ \\
        Qwen3VL-4B-Thinking & \checkmark & \checkmark & \checkmark & $\mathbf{72.76}$ & $\mathbf{81.06}$ \\
        \bottomrule
    \end{tabular}}
    \label{tab:si_ablation}
\end{table}

\paragraph{Comparing with Direct SFT on an Instruct Model.}
To assess whether our two-stage recipe can be replaced by directly fine-tuning a larger Instruct model, we compare against Qwen3VL-8B-Instruct trained with SFT on the answer labels. While SFT alone on the 4B Thinking model performs slightly worse than the 8B baseline ($64.16\%$ vs.\ $64.92\%$), adding the RL stage pushes the smaller model to $72.76\%$ overall accuracy, clearly surpassing the 8B Instruct model. Reasoning-based training therefore provides gains that cannot be recovered by model scale alone.

\paragraph{Effect of Color-Permutation Augmentation.}
As shown in Tab.~\ref{tab:si_ablation}, perturbing the cluster-to-color mapping during training consistently improves performance across both training stages. At the SFT stage, it lifts overall accuracy from $62.39\%$ to $64.16\%$ and known accuracy from $67.98\%$ to $70.41\%$. The gains persist after RL, with overall accuracy improving from $71.37\%$ to $72.76\%$ and known accuracy from $79.56\%$ to $81.06\%$. These consistent improvements indicate that the augmentation prevents the model from exploiting spurious shortcuts tied to specific color assignments, encouraging it to genuinely reason about character identity from visual and dialogue cues.

\begin{table}[th]
    \centering
    \caption{\textbf{Ablation of error-correction mechanisms.} We start from our full model and disable one mechanism at a time, measuring the impact on mTIoUs and PCR@Ks.}
    \setlength{\tabcolsep}{8pt}
    \resizebox{0.95\linewidth}{!}{%
    \begin{tabular}{lccccc}
        \toprule
        \tabtitle{Variant} & \tabtitle{mTIoU$_v$} & \tabtitle{mTIoU$_n$} & \tabtitle{PCR@5} & \tabtitle{PCR@10}  & \tabtitle{PCR@15} \\
        \midrule
        {Ours (Full)} & $84.90$ & $\mathbf{54.40}$ & $\mathbf{63.00}$ & $\mathbf{54.54}$ & $\mathbf{48.19}$  \\
        \midrule
        w/o Face-Track Filtering         & $83.30$ & $52.27$ & $60.00$ & $53.01$ & $47.47$\\
        w/o Query Name Filtering     & $\mathbf{85.13}$ & $52.16$ & $57.00$ & $49.46$ & $44.99$ \\
        w/o Inter-Entry Name Disambiguation & $84.90$ & $52.40$ & $62.00$ & $53.00$ & $47.48$\\
        \bottomrule
    \end{tabular}}
    \label{tab:error_correction_ablation}
\end{table}

\paragraph{Error-Correction Mechanisms.}
Tab.~\ref{tab:error_correction_ablation} summarises the contribution of each error-correction stage. Removing face-track filtering degrades both $\mathrm{mTIoU}_v$ (from $84.90$ to $83.30$) and $\mathrm{mTIoU}_n$ (from $54.40$ to $52.27$), suggesting that noisy or unreliable tracks introduce low-quality face features that weaken face clustering and subsequently affect name assignment. Removing query name filtering does not reduce $\mathrm{mTIoU}_v$ (from $84.90$ to $85.13$), but substantially reduces $\mathrm{mTIoU}_n$ (from $54.40$ to $52.16$) and $\mathrm{PCR}@10$ (from $54.54$ to $49.46$). This indicates that off-screen mentions mainly contribute spurious candidate names, which interfere with the MLLM's name-binding step rather than character localisation. Finally, removing inter-entry name disambiguation primarily affects name assignment: $\mathrm{mTIoU}_n$ decreases from $54.40$ to $52.40$, while $\mathrm{mTIoU}_v$ remains unchanged and $\mathrm{PCR}@K$ decreases slightly (e.g., $\mathrm{PCR}@5$: $63.00 \to 62.00$). This shows that inter-entry disambiguation improves the consistency of name assignments across entries without changing localising characters' first occurrences. Overall, the ablation distinguishes two input-stage filtering mechanisms that suppress noisy evidence before name binding from an inter-entry resolution mechanism that refines the final name assignments.

\section{Additional Baseline for Streaming Character Bank Construction}
\label{supp:mil_baseline}
In Sec.~\ref{exp_linkage}, we use Gemini-3-Flash-Preview and Gemini-3.1-Pro-Preview~\cite{gemini3} as baselines for streaming character bank construction. We note that these baselines are disadvantaged by our requirement that they emit face exemplars, a form of output they are not designed to produce. No existing method targets this setting directly: prior work either presupposes a pre-built character bank or assumes a recurring television cast. Proprietary MLLMs are therefore the closest available point of comparison, notwithstanding the mismatch in output format. The lenient scoring criterion was adopted to compensate that.

To provide a comparison that does not rely on lenient scoring to compensate for the output-format mismatch, we have additionally implemented a classical weakly-supervised baseline~\cite{haurilet2016naming}. It uses exactly the same face tracking, clustering, and majority-vote name binding as our pipeline, replacing only the mechanism by which names are linked to faces, and is given access to the full film rather than operating in a streaming pass. The weakly-supervised baseline uses subtitle text alone, with no audio, no pixels, and no vision-language model. A character vocabulary is extracted from the transcript by NER, and each mention is typed as first-, second-, or third-person using the five grammar rules in~\cite{haurilet2016naming}, so that first- and second-person mentions ("I'm Harry", "Hello, Harry") imply presence while third-person mentions ("Where is Harry?") imply absence. Each typed mention yields a positive or negative bag of face tracks from a surrounding window, here three shots centred on the mention's shot. A per-character linear MIL model (Normalized Set Kernel, the best-performing solver in~\cite{haurilet2016naming}) is trained on InsightFace embeddings, and a second-stage classifier, trained on each model's most confident tracks, scores every track in the film and assigns the argmax name.

\begin{table}[t]
    \centering
    \caption{\textbf{Results of the MIL baseline on the MCN First Occurrence dataset.} We report an additional baseline for streaming character bank construction.}
    \vspace{-1mm}
    \setlength{\tabcolsep}{6pt}
    \resizebox{0.93\linewidth}{!}{%
    \begin{tabular}{lcccccc}
        \toprule
        \tabtitle{Method} & \tabtitle{Streaming} & \tabtitle{mTIoU$_v$} & \tabtitle{mTIoU$_n$} & \tabtitle{PCR@5} & \tabtitle{PCR@10}  & \tabtitle{PCR@15} \\
        \midrule
        Gemini-3-Flash-Preview   & $\times$   & $51.81$ & $30.92$ & $35.0$ & $32.0$ & $29.8$ \\
        Gemini-3.1-Pro-Preview   & $\times$   & $49.05$ & $30.10$ & $44.0$ & $37.2$ & $34.9$\\
        MIL~\cite{haurilet2016naming} & $\times$   & $82.85$ & $33.38$ & $34.0$ & $19.7$ & $16.7$\\
        \midrule
        \textbf{Ours} & \checkmark & $\mathbf{84.90}$ & $\mathbf{54.40}$ & $\mathbf{63.0}$ & $\mathbf{54.5}$ & $\mathbf{48.2}$ \\
        \bottomrule
    \end{tabular}}
    \vspace{-2mm}
    \label{tab:mil_results}
\end{table}

In Tab.~\ref{tab:mil_results}, since the baseline shares our face tracking and clustering, it attains a comparable $\mathrm{mTIoU}_v$ (82.85 against 84.90), confirming that the two systems recover visual identities equally well. On the naming metrics, however, it falls far behind despite full-movie access, reaching 33.38 $\mathrm{mTIoU}_n$ against our 54.40 and 34.0 $\mathrm{PCR@5}$ against our 63.0. Dialogue cues under classical weak supervision are therefore insufficient to name characters reliably at feature-length scale, which is the gap our fine-tuned naming model closes.

\section{Details on Dataset Collection}
\label{supp:alignment}
MovieNet \citep{huang2020movienet} provides body bounding-box annotations on sparse 240P keyframes indexed by $(s, k)$, a shot index and within-shot image index. These indices do not identify frames in the native-resolution movie file, and MovieNet's shot segmentation does not coincide with ours. To build the MCN dataset, we must (i) recover the exact frame number $f^{\star}\!\in\!\{0,\dots,N{-}1\}$ in the full video for each annotated keyframe, and (ii) rescale every box from 240P to native resolution. We address this with a three-stage coarse-to-fine pipeline: CLIP retrieval for coarse localisation, longest-increasing-subsequence (LIS) filtering to remove inconsistent matches, and perceptual hashing for exact frame recovery.

\textbf{CLIP Feature Extraction.}
For every MovieNet keyframe and for every 24th frame ($\approx$\,1\,fps at 24\,fps) of each full-resolution video, we extract a 512-D CLIP ViT-B/32 image embedding. The keyframes form a \emph{query} pool $\mathcal{Q}$ and the video samples a \emph{source} pool $\mathcal{S}$. Both pools are cached per movie so that the subsequent stages run independently per title.

\textbf{Coarse Alignment via FAISS Retrieval and LIS Filtering.}
Within each movie we run exact L2 nearest-neighbor search (FAISS \texttt{IndexFlatL2}) of every query against $\mathcal{S}$ and take the top-1 source frame as a coarse anchor $\widehat{f}_i$ for keyframe~$i$. CLIP retrieval is robust to the resolution gap and to codec and color differences between annotation crops and rendered video frames, but it occasionally returns a visually similar shot from elsewhere in the movie. We prune such errors by exploiting the fact that MovieNet keyframes are listed in temporal order: the sequence $(\widehat{f}_i)_i$ must be monotonically increasing. We compute its longest increasing subsequence in $O(n\log n)$ and mark every keyframe outside the LIS as unreliable ($\widehat{f}_i \!\leftarrow\! -1$). The criterion is strict but conservative: any rogue match must drop out of the LIS together with whichever neighbor it conflicts with, and the LIS retains the longer consistent chain.

\textbf{Fine Alignment via Perceptual Hashing.}
The coarse stage anchors each keyframe to within $\pm 24$ frames. To recover the exact frame we scan a window $[\widehat{f}_i\!-\!W,\,\widehat{f}_i\!+\!W]$ of the full video with $W{=}150$. For each candidate frame and for the query we compute two 64-bit perceptual hashes on the CLAHE-normalized grayscale image: \emph{pHash} (low-frequency DCT coefficients thresholded against their median) and \emph{dHash} (horizontal gradient sign). Candidates are ranked by the combined Hamming distance $H_{\text{p}}{+}H_{\text{d}}$, and we accept the lowest-ranked frame satisfying both $H_{\text{p}}{\le}12$ and $H_{\text{d}}{\le}10$. If no candidate meets both thresholds we fall back to the lowest combined-distance frame in the window. An optional ORB+RANSAC inlier rerank on the top-$M$ candidates disambiguates near-duplicates within the same shot. When a keyframe's coarse anchor was discarded by the LIS filter, we replace its window with one bracketed by the nearest two LIS-passing neighbors in the same movie, $f_{\text{prev}}$ and $f_{\text{next}}$: we use $[\min(f_{\text{prev}},f_{\text{next}}),\,\max(f_{\text{prev}},f_{\text{next}})]$ when both exist, a $\pm W$ window around the single available neighbor when only one exists, and drop the keyframe entirely when neither exists.

\textbf{Bounding Box Rescaling and Output.}
With the exact frame index $f_i^{\star}$ recovered, each MovieNet body box is rescaled from the per-row annotation resolution $(W_q, H_q)$ to native video resolution $(W_f, H_f)$ via axis-independent scale factors $s_x{=}W_f/W_q$ and $s_y{=}H_f/H_q$. The resulting aligned records, of the form $(s, k, f_i^{\star}, \mathbf{b}_i, \mathrm{pid})$ where $\mathbf{b}_i$ is the rescaled body box and $\mathrm{pid}$ is the MovieNet character identifier, serve as the automatic ground truth for the MCN dataset.

\section{Dataset Examples}
\label{supp:dataset}
\begin{figure}[h]
    \centering
    \includegraphics[width=\linewidth]{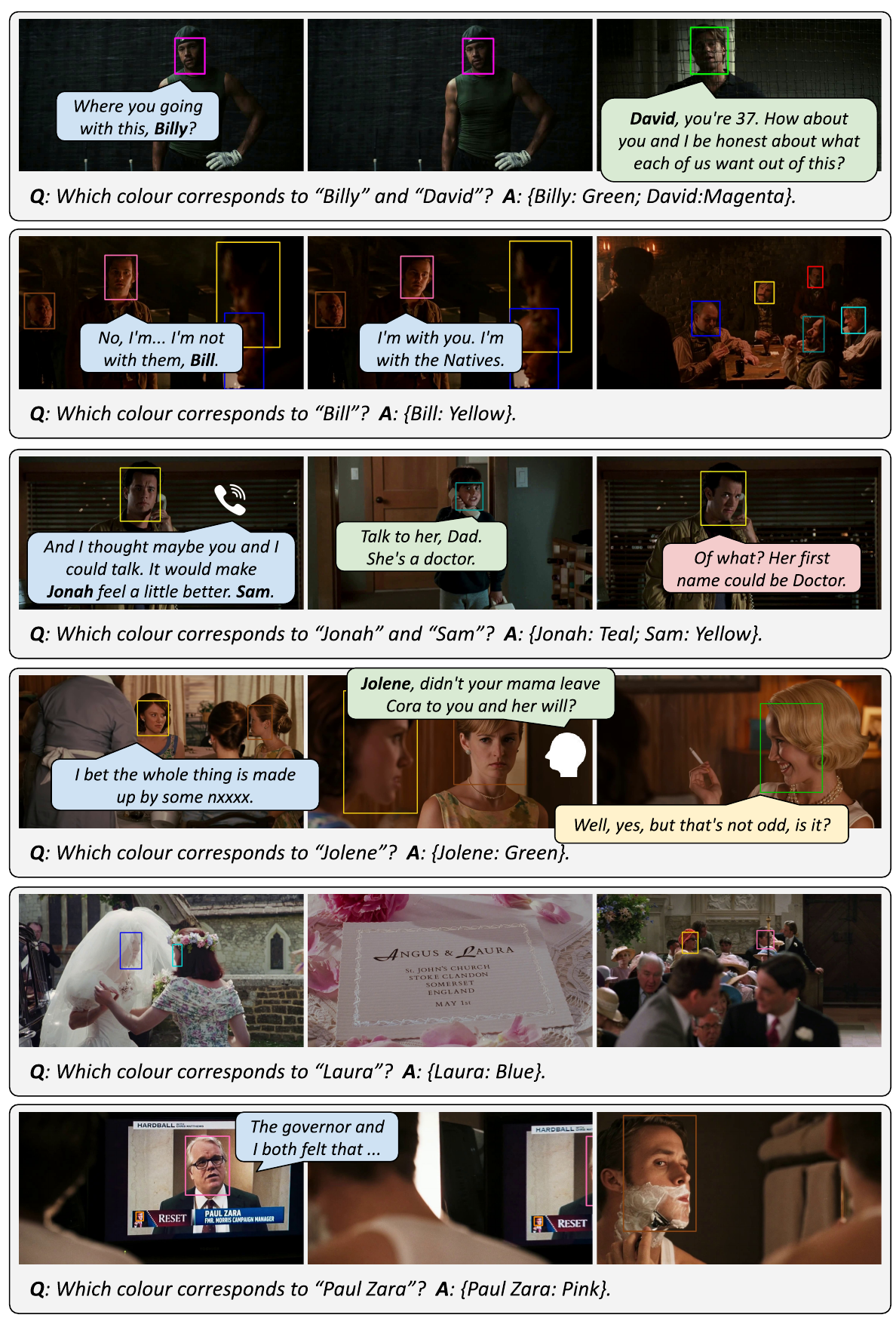}
       \caption{\textbf{Examples from the MCN dataset.} Six character-naming annotations illustrating the range of problem types covered: direct address with shot--reverse-shot structure, direct address with eyeline match, direct address over the telephone, on-screen text, and OCR-based reasoning. Clips are drawn from \textit{Moneyball}, \textit{Gangs of New York}, \textit{Sleepless in Seattle}, \textit{The Help}, \textit{Four Weddings and a Funeral}, and \textit{The Ides of March}.}
    \label{fig:dataset_visualization}
\end{figure}

Fig.~\ref{fig:dataset_visualization} shows six representative annotations from the MCN dataset, drawn from films that span genres, eras, and visual styles --- from the period crowds of \textit{Gangs of New York} to the romantic comedy of \textit{Sleepless in Seattle}, the political drama of \textit{The Ides of March}, and the sports biopic \textit{Moneyball}. This diversity is by design: a character-naming benchmark that generalizes beyond a single genre or production style must capture the full range of cinematographic conventions through which characters are introduced.

The examples also illustrate the variety of cues a model must exploit to resolve a name to the correct face. The most common case is \emph{direct address}, in which one character names another in dialogue, but the visual realisation of this act is far from uniform. The named character may be identified through the canonical \emph{shot--reverse-shot} structure, through an \emph{eyeline match} when speaker and addressee share the frame, or through a \emph{telephone conversation} where the named character is entirely off-screen at the moment of naming and must be linked to a face seen elsewhere in the clip. A separate class of examples requires reasoning from \emph{on-screen text}: the character's name appears as a written cue --- a name tag, a letter, a sign, or a screen --- and the model must perform OCR and then ground the recovered string to the correct visual identity. Cases like these are unsolvable from dialogue alone and require multimodal reasoning over both spoken and written language alongside visual evidence.

\section{Qualitative Results for Character Naming}
\label{supp:cn_example}
\begin{figure}[h]
    \centering
    \includegraphics[width=\linewidth]{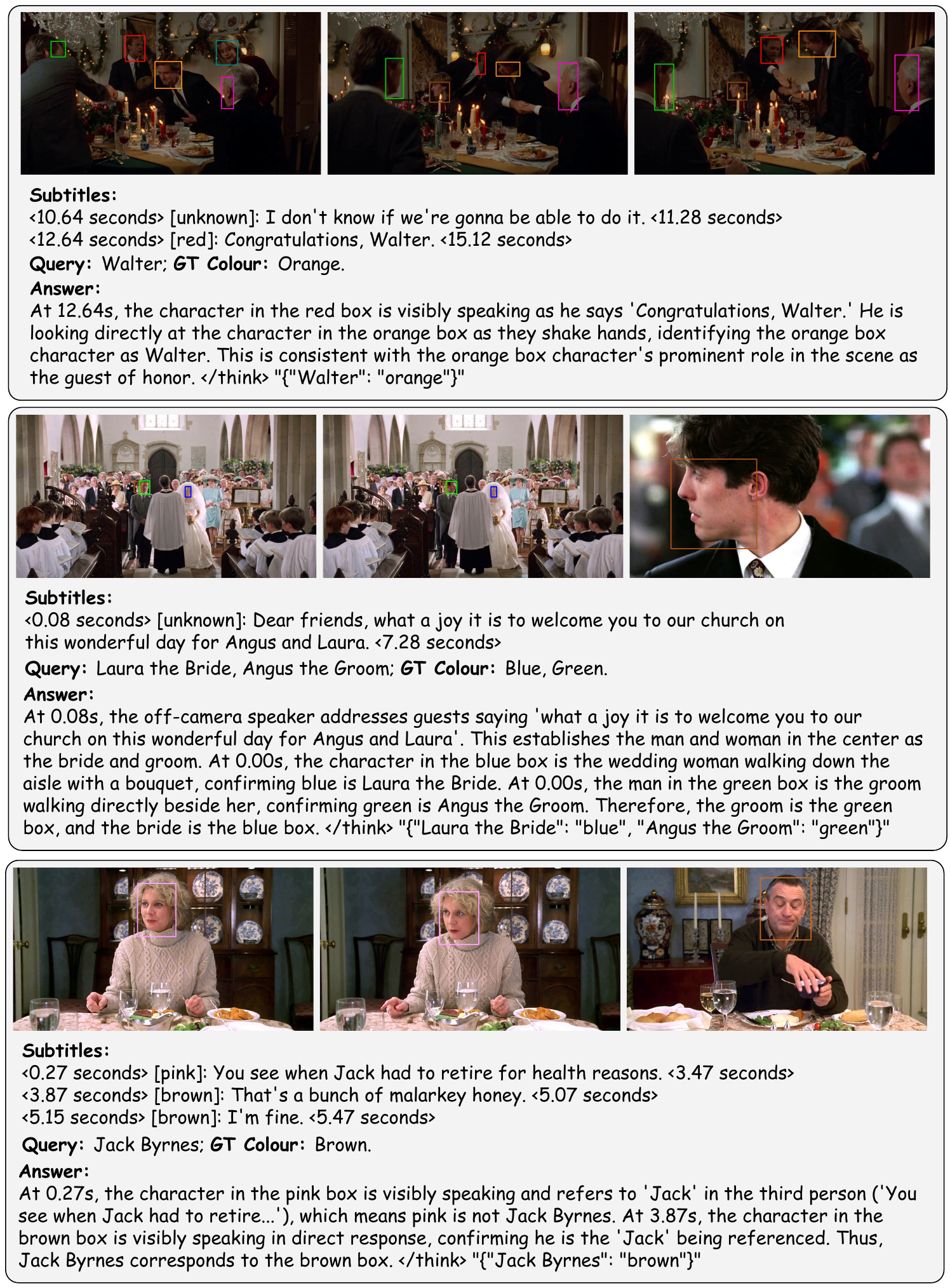}
    \caption{\textbf{Qualitative results for character naming.}
    We associate character names with their facial appearances in movies, mirroring how audiences infer who is who while watching. Characters may be introduced through diverse cues, including \textit{direct address by other characters} (top), \textit{visual reasoning with common knowledge} (middle), and \textit{third-party reference} (bottom). In each case, the model grounds the name to the correct colored box by combining the spoken cue with visual evidence such as gaze, gesture, and scene role. Correctly linking character identities is central to movie understanding, enabling coherent character tracking across scenes. The examples are taken from \textit{Sleepless in Seattle}, \textit{Four Weddings and a Funeral} and \textit{Meet the Parents}.}
    \label{fig:nia_visualization}
\end{figure}

\textbf{Qualitative Examples.} Fig.~\ref{fig:nia_visualization} shows three representative examples covering the main cue types our model handles: \textit{direct address}, where a speaker calls another character by name; \textit{third-party reference}, where a name is mentioned about someone else in the scene; and \textit{visual reasoning}, where the spoken cue is ambiguous and the mapping must be inferred from scene role. In each case, our model grounds the name to the correct colored box by jointly leveraging the spoken cue and the surrounding visual context.

\begin{figure}[h]
    \centering
    \includegraphics[width=\linewidth]{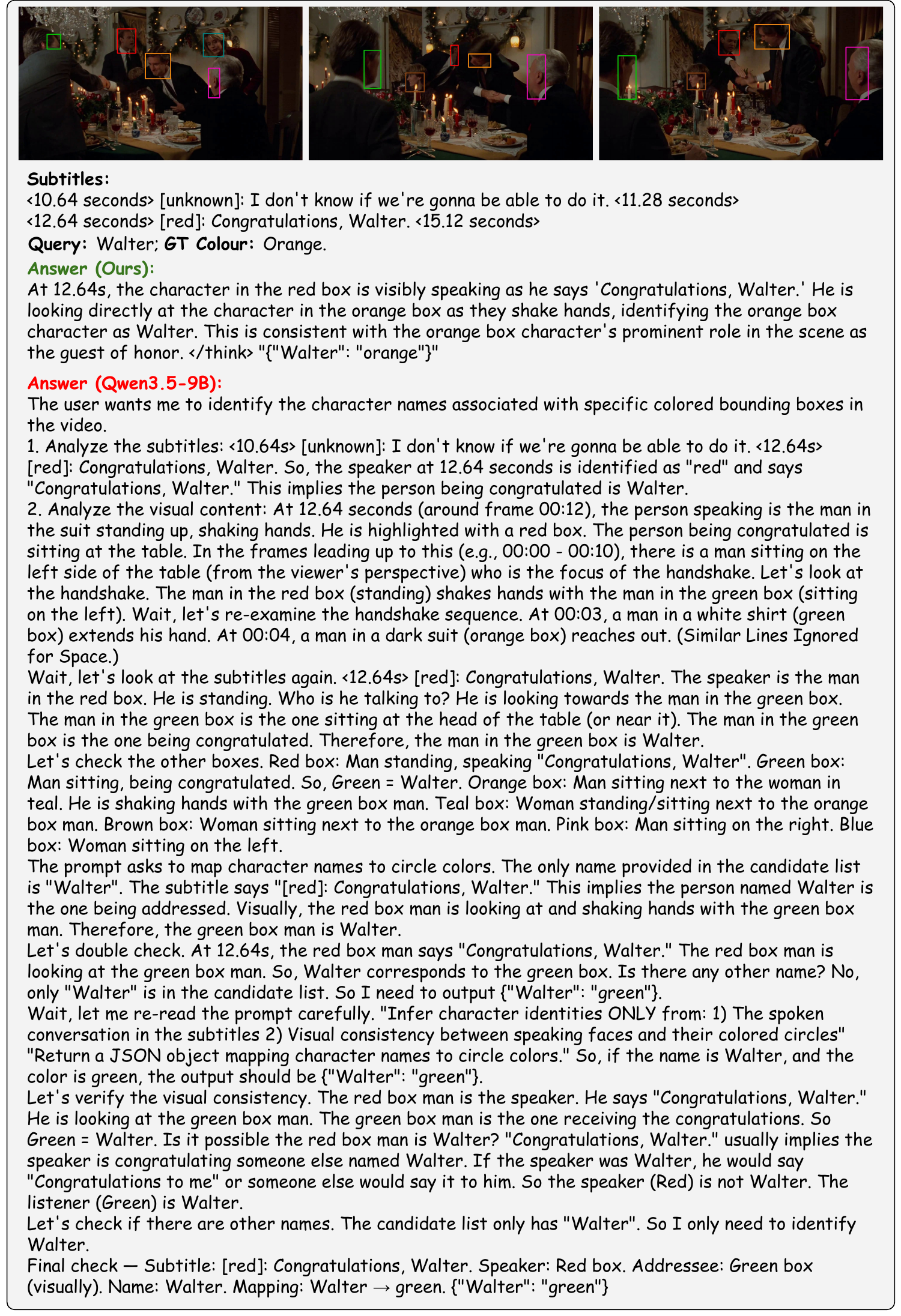}
    \vspace{-6mm}
    \caption{\textbf{Comparison of reasoning efficiency on character naming.} We compare our model against Qwen3.5-9B on the same scene. Our model produces a concise, well-grounded answer that directly links the spoken cue (``Congratulations, Walter'') to visual evidence (gaze and handshake direction), correctly identifying the orange box as Walter. In contrast, the baseline generates a lengthy chain of thought with repeated re-examinations and self-verifications, yet arrives at an incorrect mapping (green). This illustrates that longer reasoning does not guarantee better grounding: our model achieves more accurate character naming with substantially fewer tokens.}
    \label{fig:nia_comparisons}
\end{figure}

\textbf{Comparing with Baseline.} We compare our output against a strong baseline on the same scene in Fig.~\ref{fig:nia_comparisons}. Our model produces a concise, well-grounded explanation and arrives at the correct identity, whereas the baseline generates a long chain of thought with repeated self-verifications yet commits to an incorrect mapping. Longer reasoning traces do not guarantee better grounding: our model achieves more accurate association with substantially fewer tokens.

\begin{figure}[h]
    \centering
    \includegraphics[width=0.9\linewidth]{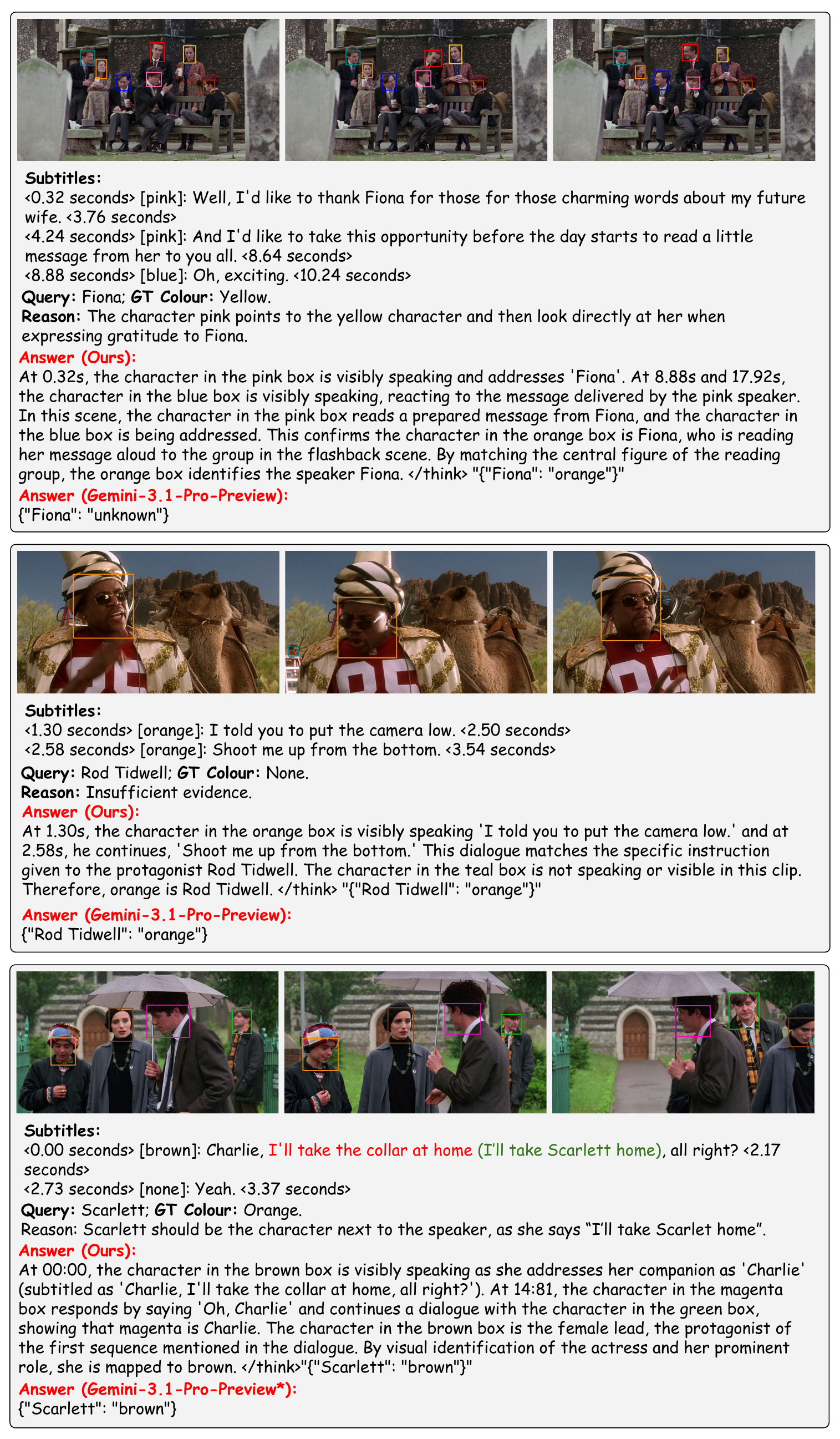}
    \vspace{-3mm}
    \caption{\textbf{Examples of failure cases for both our model and Gemini-3.1-Pro-Preview.}
    We show three examples in which Gemini-3.1-Pro-Preview, the strongest proprietary baseline, fails. These cases suggest that it struggles to capture subtle actions that are important for character naming. In the last example, Gemini is given the original audio rather than the erroneous subtitles for inference. The examples are taken from \textit{Four Weddings and a Funeral} and \textit{Jerry Maguire}.}
    \label{fig:nia_failure_cases_1}
\end{figure}

\begin{figure}[h]
    \centering
    \includegraphics[width=0.9\linewidth]{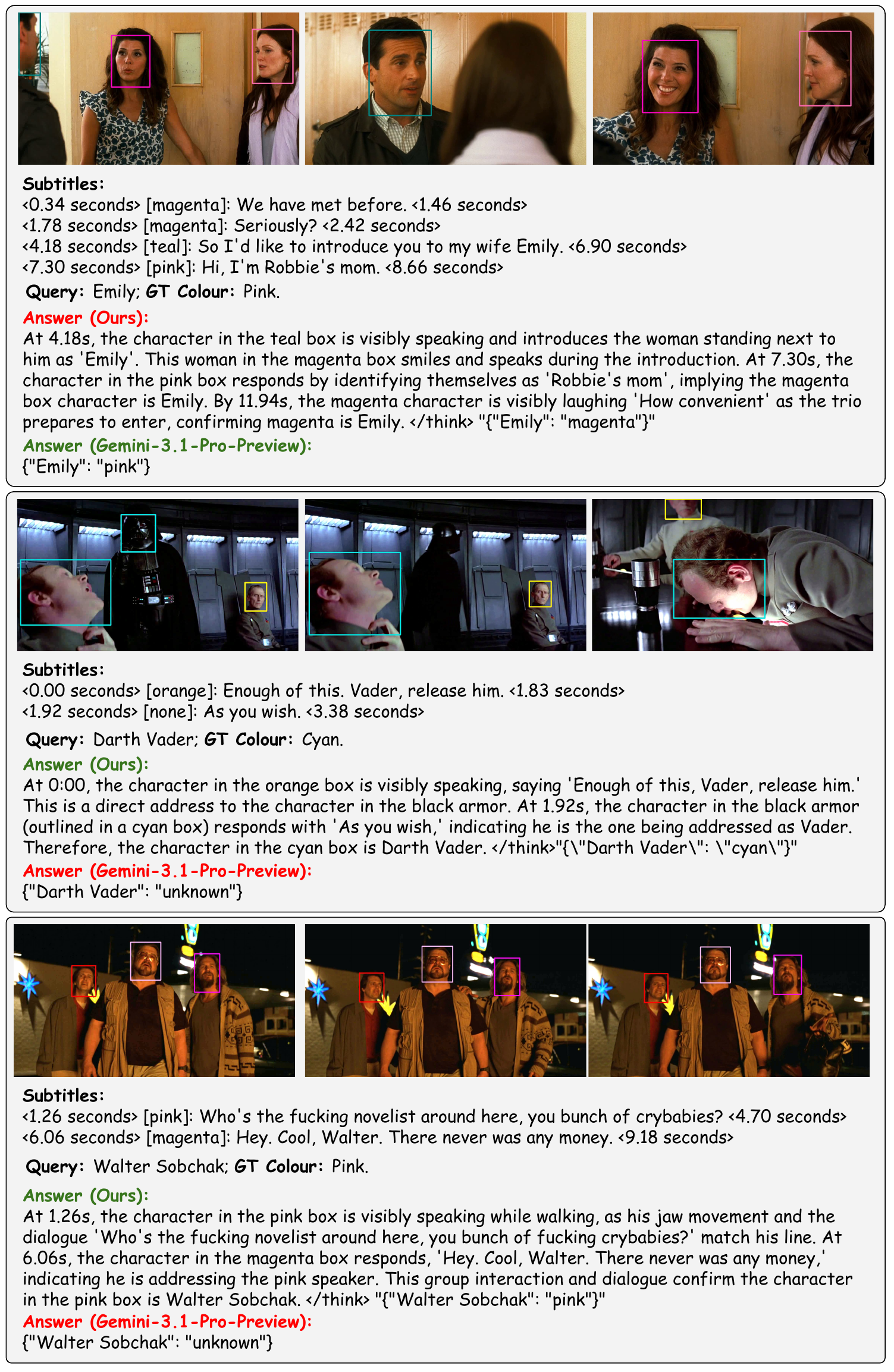}
    \caption{\textbf{Examples of failure cases for either our model or Gemini-3.1-Pro-Preview.} We show three examples in which Gemini-3.1-Pro-Preview or our model fails under common failure modes. These cases suggest that our model is better at leveraging fine-grained visual cues, speaker activity, and cross-turn dialogue context to resolve character names, while remaining errors often arise from subtle actions, ambiguous speaker grounding, or noisy subtitles. The examples are taken from \textit{Crazy, Stupid, Love.}, \textit{Star Wars: Episode IV - A New Hope}, and \textit{The Big Lebowski}.}
    \label{fig:nia_failure_cases_2}
\end{figure}

\textbf{Failure Cases.}
We analyse representative failure cases of our model and Gemini-3.1-Pro-Preview in Fig.~\ref{fig:nia_failure_cases_1} and Fig.~\ref{fig:nia_failure_cases_2}. Fig.~\ref{fig:nia_failure_cases_1} shows cases where Gemini-3.1-Pro-Preview fails despite being the strongest proprietary baseline, suggesting that it can struggle to use subtle visual actions and interactions that are crucial for character naming. Fig.~\ref{fig:nia_failure_cases_2} further compares common failure modes across both models. Overall, our model appears better at leveraging fine-grained visual cues, speaker activity, and cross-turn dialogue context, while remaining errors are often caused by subtle actions, ambiguous speaker grounding, or noisy subtitles.

\section{Qualitative Results for Streaming Character Bank Construction}
\label{supp:qual_scbc}
\begin{figure}[h]
    \centering
    \includegraphics[width=\linewidth]{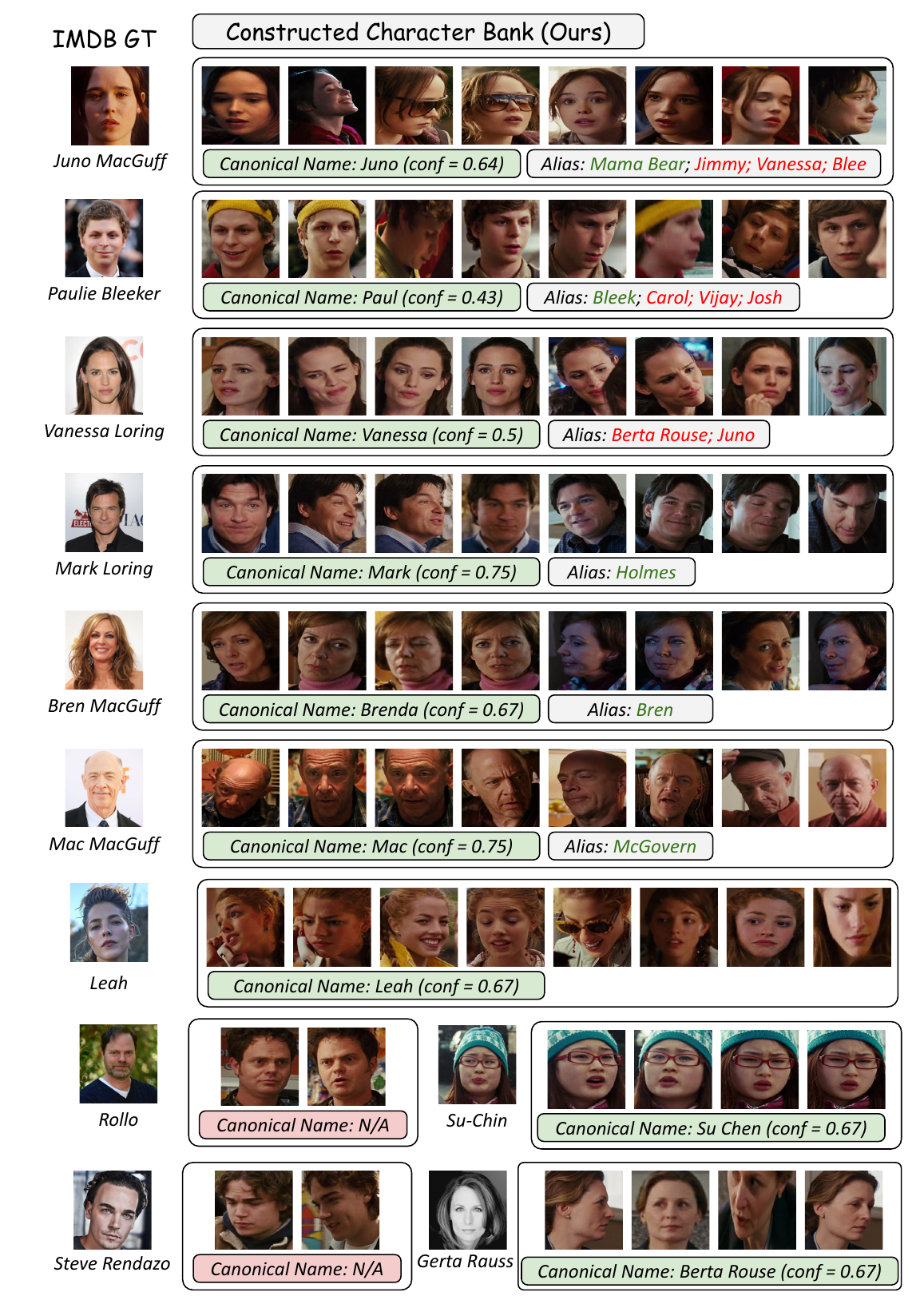}
    \caption{\textbf{Example of the constructed character bank for \textit{Juno}.}
    We show the character bank constructed by our streaming pipeline. Each row (apart from the last two)
    corresponds to one character entry and includes representative face tracks, the predicted canonical name with confidence, and discovered aliases. The example illustrates that the bank captures major characters with clean visual clusters and resolves name variations, such as mapping \textit{Mama Bear} to \textit{Juno}, \textit{Bren} to \textit{Brenda}, and \textit{McGovern} to \textit{Mac}. Some naming errors remain, typically due to noisy or ambiguous character-name associations.}
    \label{fig:Juno_visualization}
\end{figure}

\begin{figure}[h]
    \centering
    \includegraphics[width=\linewidth]{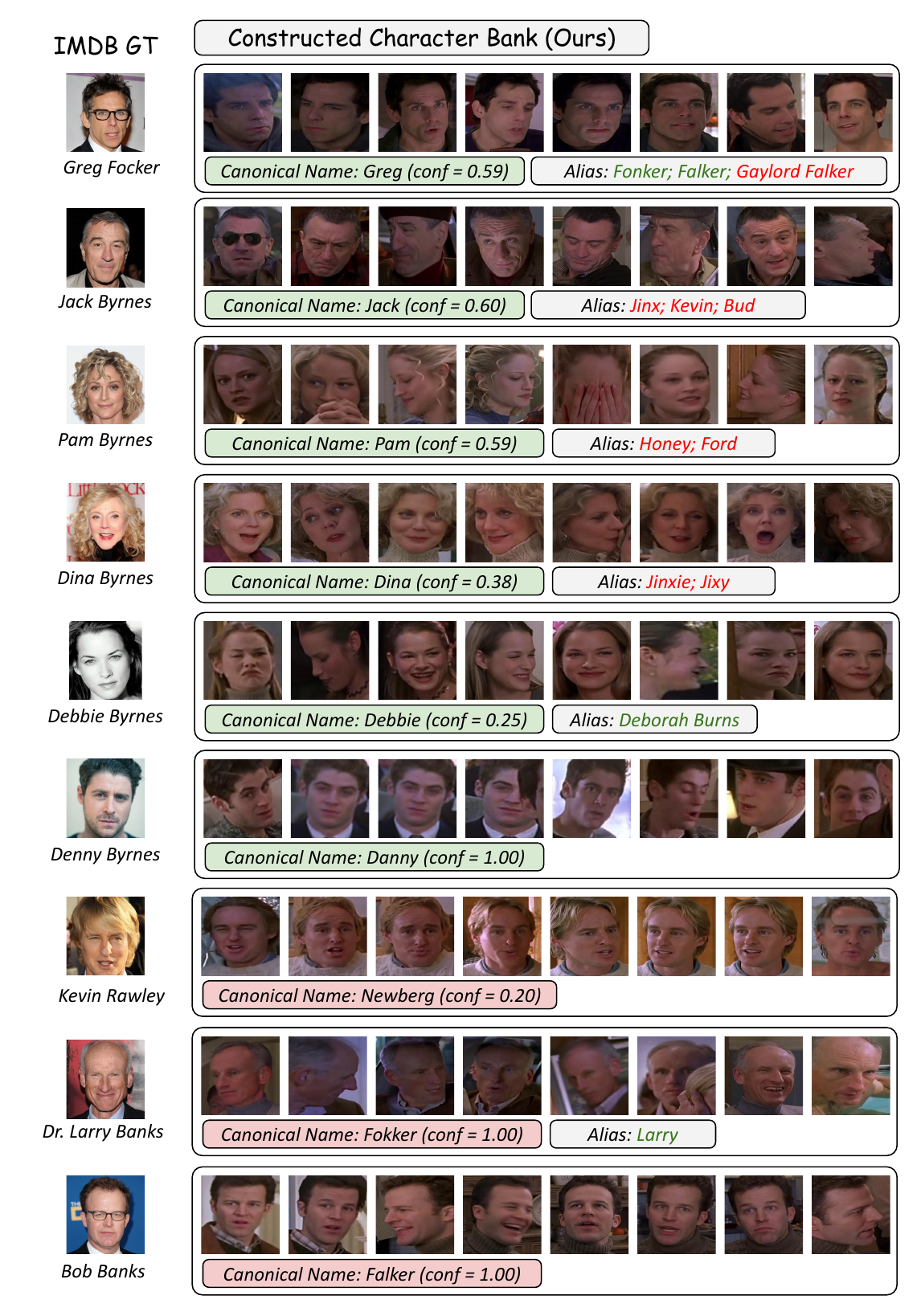}
    \vspace{-3mm}
       \caption{\textbf{Example of the constructed character bank for \textit{Meet the Parents}.}
    We show another example of a character bank constructed by our streaming pipeline. Our method correctly identifies the most prominent characters, but errors remain for the bottom three entries. These characters often appear in complex multi-person scenes, making reliable name binding more difficult. In addition, transcription errors can introduce noisy name variants; for example, \textit{Fonker} and \textit{Falker} should both be resolved to \textit{Focker} for \textit{Greg Focker}.}
    \label{fig:Meet_the_parents_visualization}
\end{figure}

We present qualitative results for streaming character bank construction on two films, \textit{Juno} and \textit{Meet the Parents}. As shown in Fig.~\ref{fig:Juno_visualization}, our pipeline constructs clean character entries for the major characters in \textit{Juno}, grouping representative face tracks and linking them to canonical names while also preserving discovered aliases. Fig.~\ref{fig:Meet_the_parents_visualization} shows a second example on \textit{Meet the Parents}. The bank again identifies the most prominent characters, but also reveals typical failure cases in complex multi-person scenes, where ambiguous visual context and transcription errors can lead to incorrect or noisy name bindings. These examples demonstrate that the proposed bank construction procedure is effective across different films, while remaining errors are primarily due to challenging name evidence rather than face-track clustering. \emph{Moreover, we refer readers to the supplementary video, which shows the full streaming character bank construction process for \textit{Juno}.}

\section{Experiment Compute Requirements}
\label{supp:compute}
Fine-tuning of the MLLM is performed using four NVIDIA Ada 6000, A40, or L40S GPUs, each with 48 GB of memory. All inference on the MCN character-naming test set is conducted on a single NVIDIA Ada 6000, A40, or L40S GPU with 48 GB of memory. Construction of the streaming character bank also uses four NVIDIA Ada 6000, A40, or L40S GPUs.

\section{Broader Impacts}
\label{supp:broader_impacts}
Our method identifies characters in movies using cinematic and dialogue cues. This technique can be used to generate accurate audio descriptions and speaker-attributed subtitles, improving accessibility for audiences with visual or hearing impairments.
This method does not perform open-world face recognition or release any identity gallery from personal video sources. We therefore do not foresee significant negative societal impacts from this work.

\end{document}